# A System View of the Recognition and Interpretation of Observed Human Shape, Pose and Action

Copyright © 2015 David W. Arathorn

David W. Arathorn
Dept of Electrical and Computer Engineering
(formerly of Center for Computational Biology)
Montana State University-Bozeman
dwa@cns.montana.edu
and
General Intelligence Corporation
Bozeman, MT
dwa@giclab.com

## Abstract

There is neurophysiological evidence that our ability to interpret human pose and action from 2D visual imagery (binocular or monocular) engages the circuitry of the motor cortices as well as the visual areas of the brain. This implies that the capability of the motor cortices to solve inverse kinematics is flexible enough to apply to both motion planning as well as serving as a generative model for the visual processing of human figures, despite the differing functional requirements of the two tasks. This paper provides a computational model of the cooperation between visual and motor areas: in other words, a system view of an important class of brain computations. The model unifies the solution of the separate inverse problems involved in the task – visual transformation discovery, inverse kinematics, and adaptation to morphology variations – using several instances of the MSC (Map-seeking Circuit) algorithm. While the paper is weighted toward the exposition of a neurobiological hypothesis, from mathematical formalization of the problem to neuronal circuitry, the algorithmic expression of the solution is also a functional machine vision system for human figure recognition, and 3D pose and body morphology reconstruction from monocular, perspectiveless input imagery. With an inverse kinematic generative model capable of imposing a variety of endogenous and exogenous constraints the machine vision implementation acquires characteristics currently unique among such systems.

## Introduction

The ability to understand the three-dimensional structure of human pose and action observed in two dimensions seems a natural and effortless task, yet, due the variability of body-shape and pose, mathematically it is one of the most difficult problems in perception. It involves the concurrent solution of several ill-posed problems. Two decades ago a population of neurons in the macaque F5 premotor cortex were found to be activated selectively by both the witnessing of, and the execution of, ordinary physical actions [1 -7]. For this dual property they were dubbed *mirror neurons*. The existence of such neurons, and subsequent imaging evidence [e.g. 46] are evidence of the cooperation of visual and motor areas in what would seem to be a purely visual task. That the brain's motor areas, whose primary task is planning and/or controlling physical motion in three dimensions, is engaged by the visual system in the interpretation of two dimensional retinal imagery of body motion, provides a compelling clue to the mechanism behind the visual function. The fact that the same areas are found to be engaged when an action is performed and when the action is observed has several implications about the computational capabilities of such neuronal circuits which, to the author's knowledge, are not commented upon elsewhere. First, the inverse kinematics problem involved in generating action differs from that involved in observing and interpreting action in the following way. The planning of most actions involves specifying the positioning of the hand or foot at a desired point in space moving with a desired velocity, and then allowing a *inverse kinematics* computation to solve for the joint angles of the rest of the body to accomplish that goal. Constraints of self-obstruction and external obstacles limit the positioning of the other joints but almost always allow a range in space. In pose reconstruction, the location of each visible joint in the image constrains the position of the corresponding kinematic joint in 3-space to lie along the line of view from the joint in the image, and the method of solution must allow all these constraints to be applied. If the same motor circuits are in fact engaged in both planning and observation, those circuits must allow operation in both modes just described. On the other hand the execution of an action almost certainly involves a forward kinematic computation. If, for example, in the course of executing a reach the hand makes contact with an unexpected obstacle, the brain knows where that obstacle lies in space. That is the result of a *forward kinematic* computation: i.e. given the current joint angles, where is my hand in space? If the same circuits are involved in both execution and reconstruction, then those circuits are equally capable of inverse and forward kinematic computations. These multiple requirements are an indication that the neurons involved could be organized in a circuit with the properties of the MSC (Map-seeking Circuit) algorithm [9, 10, 11]. As of this writing there are no other algorithms with a plausible neural instantiation capable of all the required modes of computation.

The system responsible for the solution of the pose reconstruction/interpretation problem must solve both the kinematic and visual transformation components. There is another transformation required: between the kinematic "skeleton" representation of a body and the fleshed out image of that body on the retina. In the system proposed here that transformation takes place in two steps: first the kinematic skeleton is fleshed out to a canonical 3D figure, and then the shape of that canonical figure is adjusted to match the shape of the figure in the image. That shape adjustment is termed the morphing transformation.

From a computational point of view (and hence from a machine vision point of view) the system described here is functional and complete and stands on its own, as the demonstration data indicates. From a computational neuroscience point of view, the system is offered as a hypothesis for which, as of this writing, not enough hard neurophysiology or circuit-level neuroanatomy is known to either confirm or refute. Due to the theoretical orientation of this paper, and considerations of length, there is little emphasis on the engineering aspects of the software implementation of the model. Since it is based on a known algorithmic technique (MSC), much of this can be found in other publications or on-line. Less well known is the neurobiological interpretation of MSC, having been the subject of only one publication [9]. For this reason, the possible connections between the MSC algorithm and cortical circuitry, both visual and motor, are reprised in this paper. The full exposition follows a path from the mathematics of the problem, to its algorithmic solution, and then to plausible neural circuitry.

Recent research indicates that the brain encodes action as an evolution of poses [43]. Since the determining the articulation parameters of an evolution of poses becomes much simpler once the parameters of the starting pose have been determined, it is essential to explain how the brain may solve the much more difficult latter problem. Since humans can interpret images of contortions whose prior probability is virtually nil with same apparent ease as interpreting everyday poses, it is safer to start with the assumption that whatever method humans use to accomplish this interpretation doesn't appear to limit or prioritize the search space of pose solutions by the probability of each pose. This, and several other considerations characterize the chief difficulties of the problem:
(1) With no prior knowledge to restrict it, the parameter space of all physically possible poses is very large.
(2) The viewpoint from which the body is observed seems to make little difference to the ability to determine pose parameters, nor does the scale of the body image on the retina.
(3) Since pose parameters can be determined from distant, monocular observation, depth information from stereo vision or perspectivity is not essential to the process, though it may assist.
(4) The variety of morphology of the body and the additional variations from clothing (within limits) does not materially impede determination of pose parameters.
(5) Since under normal circumstances the individual has no means to observe his own actions from the same viewpoint as he observes other individual's actions, normal learning protocols are not applicable to acquiring the ability to map articulation parameters to image data. That is to say, the acquisition of this ability cannot proceed by supervised learning because there is no practical way, without a mirror, of confirming that the pose the observer strikes to imitate the pose of the observed is in fact congruent. Since the ability obviously pre-dates mirrors it is either hard-wired according to genetic specifications, or it somehow trains up unsupervised. No solution to the acquisition of the skill is proposed here: it is assumed to have been acquired by whatever means and be in an operational state.

In the machine vision literature, the term *recognition* has by usage come to imply association of the image of an object with a label. Here *recognition* implies the association of an image with a model. In the machine vision literature the inverse problem of determining kinematic pose parameters from image data is often referred to as *pose reconstruction*. We will use that term, or *pose interpretation*, here. There is a substantial machine vision literature on this problem, some of which will be considered later. The mathematical tool, MSC, applied to the problem here has also been applied successfully to other object recognition and image registration problems, most relevantly to determining the configuration of articulated mechanical objects [10, 11]. It appears for the first time here applied to the human pose reconstruction problem. Figure 1 presents an example of the problem and its solution using the computations being proposed here: an input image containing a figure in an early phase of a throwing action, and a 3D reconstruction of that pose. (Where space allows, multiple viewpoints of 3D reconstructions are presented in this paper to preclude hiding possible reconstruction errors in depth which are invisible from the viewpoint of the input image. See Figure 10.) Other pose reconstructions appear in Figures 3, 5, 10 and 17. An example of reconstruction of an action or evolution of poses is seen in Figure 2.

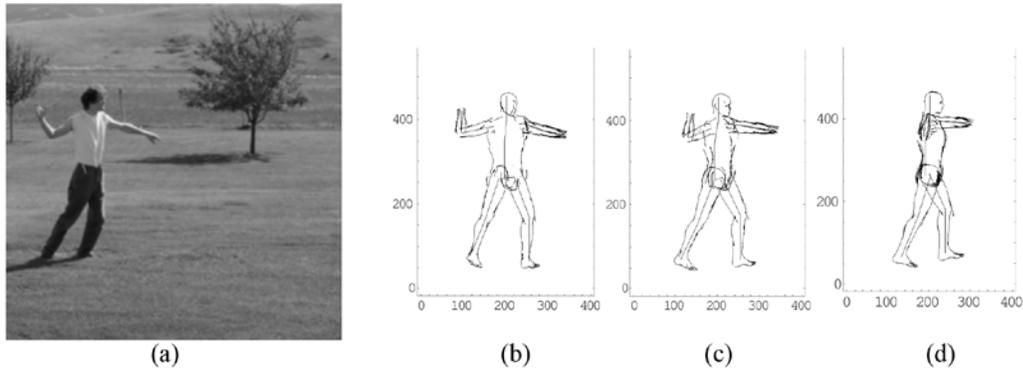

Figure 1. Recognition and pose reconstruction: (a) input image, (b-d) different views of the same 3D pose reconstruction. Multiple viewpoints of the 3D reconstruction are necessary to reveal errors in the z-axis (depth) of the input image. The determination of the presence and localization of a body, if there is one, in the input image is the *recognition* part of the problem. It is solved concurrently with the pose reconstruction in this system. The resolution of the figure in (a) is roughly equivalent to 1-2 degree eccentricity, subject at 15m.

An example of reconstruction of an action or evolution of poses is seen in Figure 2.

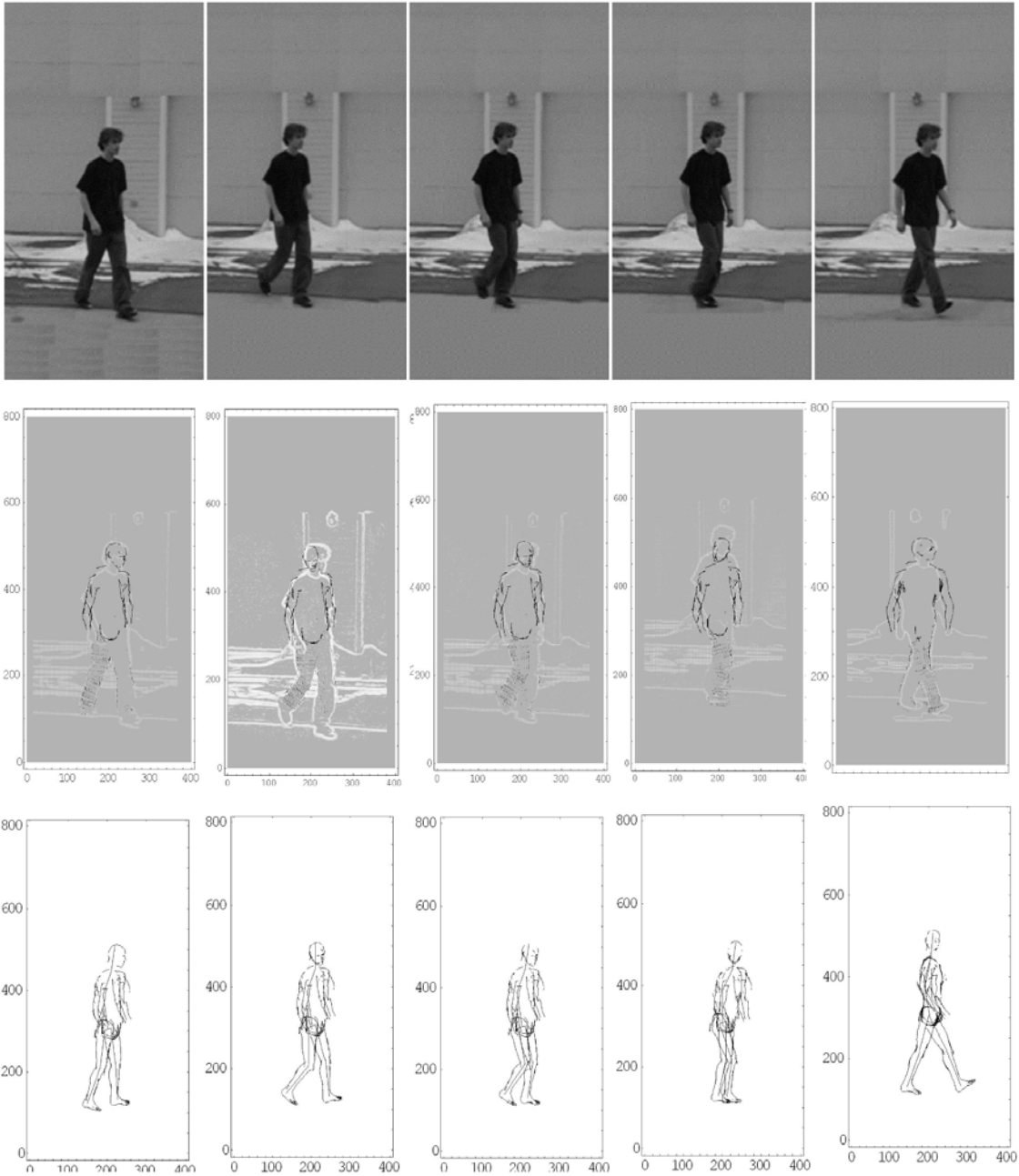

Figure 2. Reconstruction of action: (top) input image, (middle) overlay of 3D pose reconstruction, (bottom) reconstruction from different viewpoint for verification.

The body of this paper is organized as follows:
-- We identify the relationship between inverse kinematics in combination with visual transformation discovery and the pose reconstruction problem, and formalize the problem to be solved.
-- Next, we demonstrate that every aspect of the problem can be represented as a subproblem of determining the mapping from one point in 2D or 3D space to another and that the whole problem can be expressed as a composition of such mappings.

-- Next we present the algorithm that can be used to solve the problem tractably when expressed as a composition of mappings.
-- Finally we present neuronal circuitry which is the direct analog of the algorithm and is capable of solving the same problem.
This is followed by a discussion of the relationship to other models and some predictions that emerge from the model presented here.

## Method and Model

**The relationship of inverse kinematics and pose reconstruction**
The planning and execution of action involves two classes of computation: forward kinematics and inverse kinematics (IK). Forward kinematics determines the position or trajectory in space of skeletal joints given some starting reference location and each of the joint angles, or evolution of joint angles, for a chain of skeletal segments of known length. Inverse kinematics (IK) determines the joint angles, or evolution of joint angles, necessary to place the end of a particular skeletal segment at a desired location in space, or to move it along a desired trajectory in space. In typical robotic IK computations, the 3D target location for the end effector (e.g. the hand) is fully specified while the intermediate joints are free, or loosely constrained, to take whatever positions are required. In pose reconstruction all non-occluded joints are constrained to lie along lines through the locations of those joints in the image plane (Figure 3). This leaves depth of the solution with respect to the image plane indeterminate (assuming no depth cues from disparity or perspectivity). This depth indeterminacy is resolved by interposing a set of models in space onto which the image is projected. The models are articulated and morphed over the full range of possibilities allowed by the joint angle constraints and morphing limits. A matching process takes place, made highly efficient by the MSC computations, to prune the set of articulation and morphing possibilities of the model hypotheses to those consistent with the pose and shape of the body in the image. The image itself is shifted, scaled and if necessary rotated in the plane, also by MSC, to optimize the match. Some external constraints, such as gravity, may be used to reduce any resulting ambiguity in depth for limb articulations. The application of constraints is crucial to correct reconstruction, as will be discussed later.

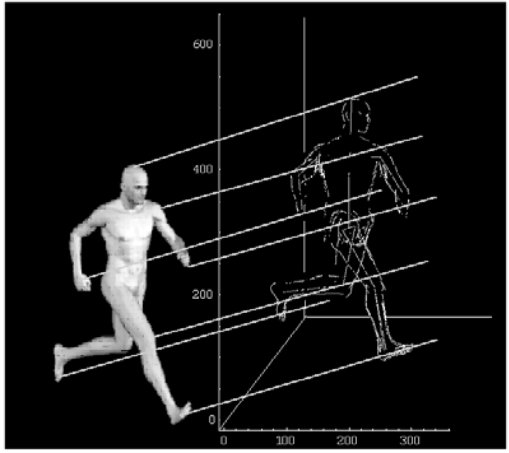

Figure 3. Indeterminacy in depth in the kinematic space.

In this problem two domains interact: the visual and the kinematic. Clearly, the representations of the two domains must be readily mappable to each other. This is readily accomplished if both domains are represented as mosaics of points in space: a pixel-like 2D space for images, and a voxel-like 3D space for kinematics. The operations on these spaces must implement visual transformations and the kinematic articulations of skeletal segment chains, respectively. And there must be one or more stages to map from the idealized "bones" of the skeletal representation to the fully fleshed and optionally clothed contours of the imaged body, with all the variation allowable in that transformation.

**Generating the skeletal articulation model using mappings**

A model of skeletal articulations can be generated by starting at some point of origin in a body-centered coordinate system, **r**, and then linking segments corresponding to the lengths of bones or groups of bones with the necessary orientations to generate any permissible pose. For example, the left upper arm kinematic range can be represented by a set of vectors $\mathbf{V}^{LUA}$ which discretize the permissible upper arm orientations relative to the shoulder

$$\mathbf{V}^{LUA} = \left(\mathbf{v}_1^{LUA}, \cdots, \mathbf{v}_n^{LUA}\right) \quad \text{where} \quad \left|\mathbf{v}_i^{LUA}\right| = \text{length upper arm segment} \tag{1}$$

Similarly for the lower arm

$$\mathbf{V}^{LLA} = \left(\mathbf{v}_1^{LLA}, \cdots, \mathbf{v}_n^{LLA}\right) \quad \text{where} \quad \left|\mathbf{v}_i^{LLA}\right| = \text{length lower arm segment} \tag{2}$$

If the shoulder socket is at location **s** in the coordinate system of the body, then all the locations reachable by the distal end of the upper arm are

$$\left(\mathbf{s} + \mathbf{v}_1^{LUA}, \cdots, \mathbf{s} + \mathbf{v}_n^{LUA}\right) \tag{3}$$

The location in space reached by the end of the lower arm in a particular orientation $\mathbf{v}_j^{LLA}$, given a particular orientation of the upper arm $\mathbf{v}_i^{LUA}$ is simply

$$\left(\mathbf{s} + \mathbf{v}_i^{LUA}\right) + \mathbf{v}_j^{LLA} \tag{4}$$

Computation of the end locations can also written as functions or mappings of a starting location **x**

$$\lambda_i^{LUA}(\mathbf{x}) = \mathbf{x} + \mathbf{v}_i^{LUA}, \quad \lambda_j^{LLA}(\mathbf{x}) = \mathbf{x} + \mathbf{v}_j^{LLA} \tag{5}$$

which allows the two segment computation in Equation 4 to be expressed as a composition of mappings applied to the shoulder joint location **s**

$$\lambda_j^{LLA} \circ \lambda_i^{LUA}(\mathbf{s}) = \lambda_j^{LLA}\left(\lambda_i^{LUA}(\mathbf{s})\right) = \left(\mathbf{s} + \mathbf{v}_i^{LUA}\right) + \mathbf{v}_j^{LLA} \tag{6}$$

The leftmost composition notation will be used throughout because it is easier to read when there are many composed mappings.

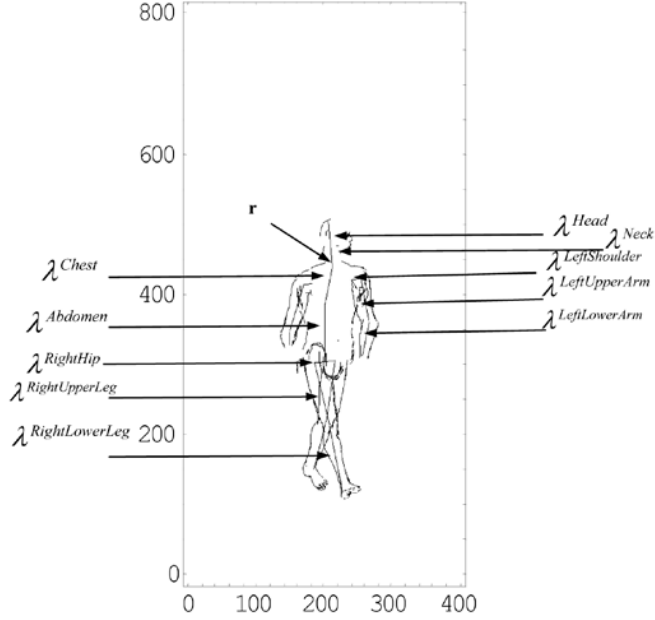

Figure 4. Generating kinematic skeleton by transformations.

The pose of the entire body can be represented as chains of composed mappings representing skeletal segments, as in Figure 4. The base of the neck is used as the point of origin **r**. The following expressions (7-12) symbolically specify the structure of the skeletal segment chains that appear in Figure 3. These will later appear as pose terms in the objective function equations which are used to solve the whole problem.

$$\lambda_{k_2}^{head} \circ \lambda_{k_1}^{neck}(\mathbf{r}) \tag{7}$$

$$\lambda_{k_8}^{leftlowerarm} \circ \lambda_{k_7}^{leftupperarm} \circ \lambda_{k_6}^{leftshoulder}(\mathbf{r}) \tag{8}$$

$$\lambda_{k_5}^{rightlowerarm} \circ \lambda_{k_4}^{rightupperarm} \circ \lambda_{k_3}^{rightshoulder}(\mathbf{r}) \tag{9}$$

Some subchains originate from the end point of another chain: e.g. the hip, upper leg, lower leg chains for left and right sides originate from the base of the spine chain, **b**

$$\mathbf{b} = \lambda_{k_{10}}^{abdomen} \circ \lambda_{k_9}^{chest}(\mathbf{r}) \tag{10}$$

$$\lambda_{k_{13}}^{rightlowerleg} \circ \lambda_{k_{12}}^{rightupperleg} \circ \lambda_{k_{11}}^{righthip}(\mathbf{b}) \tag{11}$$

$$\lambda_{k_{16}}^{leftlowerleg} \circ \lambda_{k_{15}}^{leftupperleg} \circ \lambda_{k_{14}}^{lefthip}(\mathbf{b}) \tag{12}$$

The $\lambda$s are implemented as discreet mappings at small angular intervals applied to the voxel-like discretization of 3D space. Each $\lambda$ takes every voxel in the space (or a subspace of it) to another, $p$, at a specified distance and direction. Because of the discretization, any allowable pose of the whole skeleton with $S$ segments is specified to a close approximation by a vector of articulation indices $\mathbf{k} = (k_1, \cdots, k_S)$. Each index selects a particular transformation to orient the associated skeletal segment, and computing forward from the root gives the positions in space of each segment. The resulting model can be expressed as a vector of $S$ segments, each of which is a pair giving direction and length, $d_{k_i}$, provided by transformation $\lambda_{k_i}$ and starting position, $p_{k_i}$, provided by a non-zero element in the voxelized 3D space representation.

$$\mathbf{m_k} = \left( m_{k_1}^1, \cdots, m_{kS}^S \right) \quad \text{where } m_{k_i}^i = \left( d_{k_i}, p_{k_i} \right) \tag{13}$$

In this representation, inverse kinematics computations consist of selecting which chain of articulation mappings rooted at the voxel $\rho$ places the endpoint of a particular skeletal segment in the voxel specified as the target locus, $\upsilon$.

$$\upsilon = \lambda_{k_{i+2}}^{s_{i+2}} \circ \lambda_{k_{i+1}}^{s_{i+1}} \circ \lambda_{k_i}^{s_i} (\rho) \tag{14}$$

Determining pose parameters consists of selecting which chain of articulation mappings places each un-occluded joint in some voxel, $\upsilon_j$, along the line of view through the image position of that joint.

$$\upsilon_3 = \lambda_{k_{i+2}}^{s_{i+2}} \circ \lambda_{k_{i+1}}^{s_{i+1}} \circ \lambda_{k_i}^{s_i} (\rho), \quad \upsilon_2 = \lambda_{k_{i+1}}^{s_{i+1}} \circ \lambda_{k_i}^{s_i} (\rho), \quad \upsilon_1 = \circ \lambda_{k_i}^{s_i} (\rho) \tag{15}$$

How this is done in a computationally tractable way will be described later.

**Generating visual transformations using mappings**

Because of eye motion – jitter, drift and changes of fixation – the visual system must be capable of translation of the retinal image, or the encoding of the retinal image, to assemble a single stable image. Because the image of an object on the retina varies with distance while the representation of the skeleton in the kinematic space is approximately fixed in size, a scaling operation must take place. Because the observer's head can be tipped from side to side, rotation in the viewing plane must be accommodated. The 2D image thus transformed should be mappable to a projected view of some 3D rotation of the skeleton in the kinematic space. Let $I$ be the encoding of the retinal image, and $\mathbf{m_k}$ the skeletal model, then decomposed transformations described above are notated

$$t^{2D\,rot} \circ t^{scale} \circ t^{shift}(I) \leftrightarrow t^{3D \rightarrow 2D} \circ t^{3D\,rot}(\mathbf{m_k}) \tag{16}$$

In the simulations the 3D to 2D projection is orthographic on the assumption that the viewed body is distant or viewed in several narrow angle fixations. This allows scaling alone to accommodate differences in distance. Since the 3D rotation is not known a priori, and the scale is usually only approximately known, those two transformations must also be determined as an inverse problem. Arguably vestibular inputs could select 2D rotation to compensate head orientation. However, for simplicity, we assume here that all visual transformation parameters must be determined as an inverse problem of aligning the image of a body with the kinematic skeletal representation.

Once past the surface of the retina, the image is no longer a simple representation of the illumination in the scene, but some higher level encoding which includes at least oriented edge-like features and more complex features. For this discussion the details of the representation are not important, but we assume some edge-like features. The algorithm to be used to arrive at a solution is agnostic with respect to representations: they need only be subject to the appropriate transformations and sparse.

**Mapping between image and skeleton representation**
The mapping indicated by the double arrow in (16) glosses over a significant complication: the image is fleshed out, while the kinematic skeleton is less than bones. Yet humans have a remarkable ability to recognize an individual's gait from just the motion of spot markers at their joints [12]. One possible explanation for this ability, for which there is no apparent evolutionary advantage, is that the 2D image of those markers can be directly mapped to the 3D skeletal model segment endpoints. This would suggest that the mapping between the 3D kinematic skeleton representation and the 2D representation of a body is highly flexible and not limited to learned or evolved associations.

There are two possible categories of mapping between these two domains. It is possible that some means exists of detecting the position of joints or the axes corresponding to the skeletal segments within the image of the fleshed out and possibly clothed body. This would allow a direct mapping from image to kinematic skeleton. It would also explain our ability to process stick figures and even joint marker figures. On the other hand, it would not explain how we have such an acute sense of the body volumes of an observed human figure. This ability suggests that the mapping from skeleton to observed body incorporates a transformation through a 3D surface hypothesis. The latter option is taken in the implementation used here, though the whole architecture could as well be applied to the former.

One could still argue that no current evidence resolves whether the fleshing out of the skeleton to match the image takes place in the 3D domain or in the 2D domain. Geon theory proponents [13] would support the former, while proponents of view-based theories of object recognition [14] would support notion that the projection of the skeleton provides an axis and length for an image of the relevant body segment, and one of a library of such is substituted for the skeleton segment . Alternatively, the brain may access a collection of single viewpoint 3D component models, as may be implied by [15], which would provide the surface curvature necessary to compute shading, texture gradients and disparity. Any of these can be accommodated in the framework proposed here, so arbitrarily we proceed with the assumption that the skeleton is projected to 2D and then mapped as necessary to the occluding contour of the body in the image. This process starts by using the orientation parameters of the skeletal segment to select the occluding contour of a canonical model of that segment. The occluding contour is then widened or narrowed by translation perpendicular to the segment axis, creating a "quiver" of morph variants. Each of these morph variants is then locally deformed over a narrow range, to match the body edges in the image. The use of several morph variants limits the necessary range of deformation and thereby reduces the tendency to match random edges in the background. This operation is functionally equivalent to testing a small library of images of the body segment with deformation to optimize the match of each with the image. In either case the view and the position are constrained by the skeletal model.

In the backward or top-down pathway, the morphing operation naturally combines with the 3D to 2D projection step, designated $t_j^{morph3D-2D}$. 3D rotation, projection to 2D and morphing are applied separately to each skeletal segment and the resulting 2D component images are assembled by summation in the 2D image domain as indicated in (17) and illustrated in Figure 5(c, h).

$$t_{i_a}^{2D\ rot} \circ t_h^{scale} \circ t_g^{shift}(I) \leftrightarrow \sum_{s=1,S} \left( t_{j_s}^{morph3D-2D} \circ t_{i_b}^{azim \times elev\ rot}(m_{k_s}^s) \right) \quad (17)$$

Now that the model is fleshed out by morphing, the mapping indicated by the double arrow is a congruence.

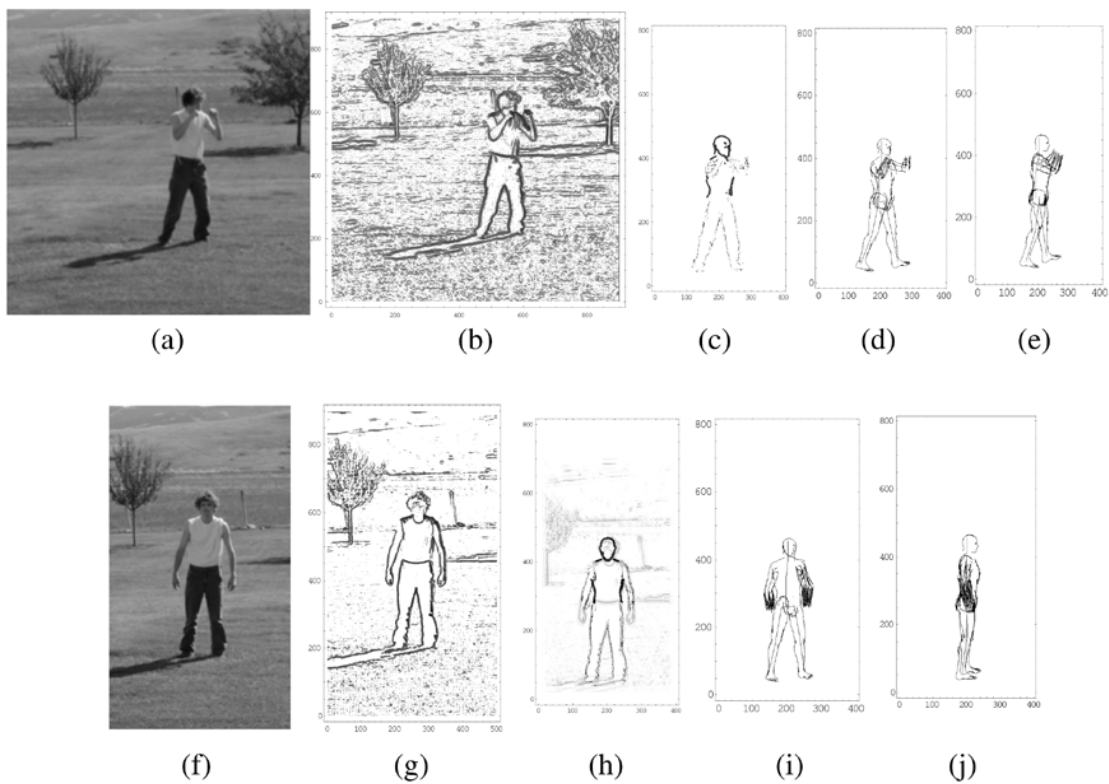

Figure 5. Stages of transformation: (a,f) input images, (b,g) edge maps, (c) morphed pose solution, (h) morphed pose solution and forward translation of edge map overlaid, (d-e, i-j) unmorphed pose solutions from different viewpoints. The orientation, pose, morphing and 2D image formation parameters of these examples are determined by the inverse method described below. Note that (h) exhibits the solution to the exact localization (sometimes called *segmentation*) and recognition part of the problem which arises as an emergent property of the determination of visual transformations.

Hidden edge removal, or occlusion of one body segment by another, can be accomplished at assembly using the depth component of the skeleton segment spatial locations, available from the projection operation in the $t_j^{morph3D-2D}$ stage and assembling the 2D image using a Z-culling assembly operator, $A^{occlusion}$.

$$t_i^{2D\ rot} \circ t_h^{scale} \circ t_h^{shift}(I) \leftrightarrow A^{occlusion}\left( t_{j_1}^{morph3D-2D} \circ t_{i_b}^{azim \times elev\ rot}(m_{k_1}^s), \cdots, t_{j_S}^{morph3D-2D} \circ t_{i_b}^{azim \times elev\ rot}(m_{k_S}^S) \right)$$
(18)

The version in (17) is used in the rest of this discussion for clarity, and in the simulations for efficiency.

Finally, some comparison operation *compare*(**a**, **b**) must be performed to produce a measure of the degree of match, $q$, between the transformed image of the body and the morphed 2D projection of the model produced by a particular set of generating parameters.

$$q = compare\left( t_{i_a}^{2D\,rot} \circ t_h^{scale} \circ t_g^{shift}(I), \sum_{s=1,S}\left( t_{j_s}^{morph3D-2D} \circ t_{i_b}^{azim \times elev\,rot}(m_{k_s}^s) \right) \right) \quad (19)$$

Since each of the transformations in the left argument, $t_i^L$, have adjoints, $t_i'^L$, if we assume the compare is an inner product, (19) can be rewritten to place all the transformations on one side.

$$q = compare\left( I, t_g'^{shift} \circ t_h'^{scale} \circ t_{i_a}'^{2D\,rot} \circ \sum_{s=1,S}\left( t_{j_s}^{morph3D-2D} \circ t_{i_b}^{azim \times elev\,rot}(m_{k_s}^s) \right) \right) \quad (20)$$

**Scale of visual, articulation and morphing transformations**

As expressed in (17) the cost of evaluating all possible compositions of transformations is the product of the number of transformations to be tested in each term of the composition. An approximate discretization of the visual transformations alone might be:

$t^{shift}$: $600^2 = 360,000$

$t^{scale}$: $10\,oct\,@\,1.05 = 140$

$t^{2D\,rot}$: $@\,1° = 360$

$t^{3D \to 2D}$: $1$

$t^{azimuth \times elevation\,rot}$: $@\,1° = 129,600$

While any algorithm which requires time or neuronal resources proportional to the product of these, $2.34 \times 10^{15}$, is intractable, an algorithm which can reach a solution using time or neuronal resources proportional to the sum, 490,100, becomes quite practical. A similar reduction applies to the kinematic problem. Assuming the orientation range of the most free-jointed limb segments is discretized to about 5000 orientations, the kinematic search is reduced from $10^{11}$ to about $(4 \times 5,000 + 4 \times 2,500 + 8 \times 200) \times 4 = 126,400$.

Morphing adds from tens to hundreds of transformations for each body segment. The complete pose reconstruction problem under an algorithm with additive scaling is therefore of order approximately 600,000.

**Applying MSC to the problem**

Expressing (20) as an optimization states the pose reconstruction problem succinctly.

$$(g, h, i_a, i_b, j_1..j_S, k_1..k_S) =$$

$$\operatorname{argmax}\left( c\left( I, t_g'^{shift} \circ t_h'^{scale} \circ t_{i_a}'^{2D\,rot} \circ \sum_{s=1,S}\left( t_{j_s}^{morph3D-2D} \circ t_{i_b}^{azim \times elev\,rot}(m_{k_s}^s) \right) \right) \right) \quad (21)$$

$c(\ )$ is some *correspondence* function which measures the similarity of the patterns which are its arguments. Argmax( ) yields the index of the transformation in each stage which maximize the correspondence or match between the image and the model. Segmentation of the input image $I$ to the region of the body is not required because optimization determines the best match between the optimal pose, morph, viewpoint, scaling and shift of the model figure to match the subregion of the retinal image containing a human figure.

Ideally, an inverse problem has an convex objective function which allows a simple gradient descent solution. But the problem in (21) is discrete and has many local maxima. And, as we have seen, testing all combinations of the transformations is not tractable, and it would remain so even if it were capable of being reduced 99% by eliminating combinations with low prior probabilities. However, problems which can be expressed as the maximization of correspondence between a pattern $I$ and a transformation of a pattern $M$ via a composition of transformations

$$\mathbf{x} = \operatorname{argmax}\left(c\left(I, t_{x_1}^1 \circ \cdots \circ t_{x_m}^m(M)\right)\right) \tag{22}$$

are amenable to solution by the MSC algorithm, with computational cost proportional to the total number of transformations that must be tested. MSC explicitly decomposes the search for the correct sequence of transformations into $m$ stages each implementing $t_1^L, \cdots, t_{n_L}^L$, $L = 1, \cdots, m$. The additive scaling property is due to the application of transformations in each stage to an aggregate of transforms from the stage before. The aggregation operation is usually superposition or another operation which does not increase dimension. These are the essential principles of the MSC algorithm and of the neuronal circuit equivalent which will be discussed later.

In the proposed architecture there are two top-level MSCs converging concurrently:
 (a) the visual circuit consisting of stages computing translation, scaling, 2D (view plane) rotation, morphing and 3D (azimuth-elevation) rotation.
 (b) the kinematic circuit consisting of sub-MSCs for each chain in Equations 7-12. The visual and kinematic MSCs influence each other's convergence by passing information between them, as described earlier.

The dataflow of MSC is bi-directional. Assume the transformation stages are numbered 1 to $m$ in the forward direction. Vector $\mathbf{f}^0$ is the forward input pattern and vector $\mathbf{b}^{m+1}$ is backward input pattern.

The MSC variant used here incorporates masks, $\boldsymbol{\mu}^L$, on each of the $\mathbf{f}$ and $\mathbf{b}$ vectors to restrict the kinematic solution space. A use of such restrictions is computation of an inverse kinematic in the presence of obstructions, a robotic example of which is seen in Figure 7, later in this paper. The masks are not used in the solution of the visual parameters, as indicated by square brackets in the pseudo code below. The pseudo code also makes explicit that the only tunable parameter, $k^L$, which governs the convergence rate, can differ for different stages. Other than $k^L$, the stages of the MSC differ only in the specific transformations executed. The single exception is described later.

for $L = m, m-1, \cdots, 2$
   for $j = 1, 2, \cdots, n_L$
     if (first iteration)    $g_j^L = 1.0$                                                               (23)
     if $(g_j^L > 0)$    $\mathbf{b}^L += g_j^L t_j'^L(\mathbf{b}^{L+1})$                                  (24)
   endfor
   $\left[\mathbf{b}^L = \boldsymbol{\mu}^{L-1} \mathbf{b}^L\right]$                                                               (25)
endfor

for $L = 1, 2, \cdots, m$
   for $j = 1, 2, \cdots, n_L$
     if $(g_j^L > 0)$
        $\tau = t_j^L(\mathbf{f}^{L-1})$                                                         (26a)
        $q_j^L = c(\mathbf{b}^{L+1}, \tau)$                                                (26b)
        $\mathbf{f}^L += g_j^L \tau$                                                         (27)
     endif
   endfor
   $\left[\mathbf{f}^L = \boldsymbol{\mu}^L \mathbf{f}^L\right]$                                                                 (28)
endfor

for $L = 1, 2, \cdots, m$
   for $j = 1, 2, \cdots, n_L$
$$\Delta g_j^L = -k^L \left(1 - \frac{q_j^L}{\max(q_1^L, \cdots q_{n_L}^L)}\right) \quad (29)$$
$$g_j^L \leftarrow \max(g_j^L + \Delta g_j^L, 0) \quad (30)$$
   endfor
endfor

Either before or during the first iteration begins, all *g*'s for all stages are set to 1.0.

Then for each iteration,
   (1) the aggregates for one direction are computed, assume backward first
   where $n_L$ is number of mappings in stage *L*. For simplicity, vector addition is used
   aggregation function in the pseudo code. Usually *L* = 1 can be skipped.

   (2) in the forward direction, each $q_j^L$ for stage *L* are computed as the correspondence between the backward aggregate from stage *L*+1 and the $j^{\text{th}}$ transformation of the forward aggregate from stage *L*-1. Each transform in the stage is computed once, and used both for computing *q* and for scaling into the aggregate for the stage.

Notice also that the only storage used is for the aggregate vector **b** for each stage in the backward direction and the **q** and **g** vectors for each stage. Not all stages need to be recomputed in the first few iterations, as will be apparent from inspection of the sequence.

(3) either after each stage's $q$'s are computed, or after the $q$'s for all stages have been computed, the $g$'s for each stage are updated via a competition function implemented by (24,25). In practice, each $g$ is set to zero when below a threshold.

This completes one iteration.

In each iteration the $g$'s corresponding to the largest $q$ in each stage is unchanged, while those corresponding to the smaller $q$'s are decreased proportionally. On the next iteration the new values of the $g$'s now reweight the contribution of each transform to the aggregates. This re-weighting then alters the $q$'s for each stage in the next iteration. The $g$'s are then updated, and so forth. Finally either (a) only one $g$ in each stage is non-zero, and the sequence of transformations specified by the non-zero $g$'s is the solution, or (b) all the $g$'s in one stage got to zero, indicating no solution.

While it can be shown that such a dynamical system must converge to a solution [14], it is known that under certain conditions of input and certain sets of transformations the correct $q$ in a given stage is not always maximal due to the aggregation of transforms. This is termed a *collusion* [9, 15, 44]. Usually the mis-ordering of $q$'s resolves itself into the correct ordering after a number of iterations prune away the contributors to the collusion. On occasion, however, the $q$ corresponding to the correct transformation will be driven down by the competition so far that it goes to zero. Of course, with the computation of $\Delta g$ as it is shown above, $g$ cannot recover. Hence the system can in fact converge to an incorrect solution, or far more likely a close but non-optimal solution. However, the probability of this decreases rapidly with an increase in the sparsity of the data and with a decrease in the number of applied transformations. As noted earlier, the only tunable parameter is $k^L$, which controls the speed of convergence. If $k^L$ is too high it can cause a collusion to result in convergence to the wrong solution, which otherwise would be pruned away. Below that point, $k^L$ has no effect on the solution, only the speed to reach it.

There has been extensive empirical experience that very low error rates can be achieved with the transformations relevant to vision and kinematics in general and to their combination in articulated object recognition [10, 11]. In addition, the mathematical properties of MSC allow its probability of convergence to a correct solution to be computed a reduced computation which captures the properties of the image, the representation and the transformations implemented in the stages [45]. In real application an ROC inflection point of >0.9 correct for <0.1 false positive is readily achievable both in theory and in practice on small target objects in highly cluttered natural scenery. As noted earlier, real human vision imposes an additional stabilization problem which involves accommodating both high frequency low amplitude translational motion of the image due to drift and jitter (up to 100 Hz) and low frequency rotational motion due to torsion (up to 1Hz). An MSC-like neuronal circuit would be applicable to this task as well, as has been demonstrated in practice by the experimental and clinical use of MSC for a number of years in removing exactly such eye-motion induced artifacts from real time retinal imagery from, and for cone-targeted stimulus delivery by, adaptive optics scanning laser opthalmascope [16, 17, 18, 19].

**MSC stages for visual transformations**

For the visual transformation MSC, $\mathbf{f}^0$ is the sparsified input image (e.g. by edge filtering) and $\mathbf{b}^{m+1}$ is the skeleton model. For the transformation sequence in (18), stage 1 shifts, stage 2 scales, stage 3 rotates in viewing plane, stage 4 transforms between 2D and 3D and morphs, stage 5 rotates in azimuth and elevation.

Henceforth, the notation $T^L(\ )$ will designate the function which computes the aggregate of all transformations $t_i^L$ in stage $L$. For the moment the aggregate can be considered to be a weighted sum of all the transforms in a given stage. This applies to both visual and kinematic transformations.

$$T^L(\mathbf{v}) = g_1^L t_1^L(\mathbf{v}) + g_2^L t_2^L(\mathbf{v}) + \cdots + g_{n_L}^L t_{n_L}^L(\mathbf{v})$$
$$\Lambda^s = g_1^s \lambda_1^s(\mathbf{v}) + g_2^s \lambda_2^s(\mathbf{v}) + \cdots + g_{n_s}^s \lambda_{n_s}^s(\mathbf{v}) \qquad (31)$$

$\Lambda^s$ in the second equation of (22) indicates an aggregate of transforms for segment $s$ and is used in the kinematic MSC. A more general aggregation operator is like an $Lp$ norm on each element of the transforms of the vector $\mathbf{v}$

$$T^L(\mathbf{v}) = \alpha_p \left( g_i^L t_i^L(\mathbf{v}) \right)_{i=1..nL} = \left( \left( g_1^L t_1^L(\mathbf{v}) \right)^p + \cdots + \left( g_{n_L}^L t_{n_L}^L(\mathbf{v}) \right)^p \right)^{1/p} \qquad (32)$$

where $\mathbf{v}$ is the vector produced by the earlier stages in the pathway. This flexibility is important to neuronal implementations whose elements have limited quasi-linear ranges.

The aggregates, with their gain coefficients, produce a continuous embedding of (18). If one carries out the appropriate algebraic manipulations, the MSC as applied to the visual transformations is seen to compute the following

$$q_g^{shift} = c\left(I, t'^{shift}_g \circ T'^{scale} \circ T'^{2D\,rot} \circ \sum \left(T_s^{morph3D-2D} \circ T^{azim \times elev\,rot}(\mathbf{m}^s)\right)\right)$$
$$q_h^{scale} = c\left(T^{shift}(I), t'^{scale}_h \circ T'^{2D\,rot} \circ \sum \left(T_s^{morph3D-2D} \circ T^{azim \times elev\,rot}(\mathbf{m}^s)\right)\right)$$
$$q_{i_a}^{2D\,rot} = c\left(T^{scale} \circ T^{shift}(I), t'^{2D\,rot}_{i_a} \circ \sum \left(T_s^{morph3D-2D} \circ T^{azim \times elev\,rot}(\mathbf{m}^s)\right)\right) \qquad (33)$$
$$q_{j_s}^{morph3D \to 2D} = $$
$$\quad c\left(T^{2D\,rot} \circ T^{scale} \circ T^{shift}(I), t_{j_s}^{morph3D-2D} \circ T^{azim \times elev\,rot}(\mathbf{m}^s)\right) \quad s = 1 \cdots S$$

From this formulation it is apparent that the $q$ associated with each transformation measures the contribution of that particular transformation to the correspondence through the aggregates of transformations in the other stages. The $q$'s above are the components of the gradient, $q_i^L = \partial Q / \partial g_i^L$, of the MSC objective function

$$Q\left(\mathbf{g}^{shift}, \cdots, \mathbf{g}^{azim \times elev\,rot}\right) = c\left(I, T'^{shift} \circ T'^{scale} \circ T'^{2D\,rot} \circ \sum \left(T_s^{morph3D-2D} \circ T^{azim \times elev\,rot}(\mathbf{m}^s)\right)\right) \qquad (34)$$

where $T^L$ is defined as in (31). Since $Q$ is not convex, simple gradient descent is not applicable. Instead, competition, as implemented by (24,25), pushes the solution in each stage toward the boundary corresponding to the largest $q$. Unlike the correspondence space in (21), $Q$ has no interior local minima so the solution will always move to a boundary, or the origin (indicating no solution), as proved in [14]. An analysis of the mathematical principle in depth appears in [15]. The mode of failure mentioned earlier moves the solution toward the wrong boundary. Due to the sparseness of skeleton segment endpoints in 3D and the occluding contours in 2D, in this application collusion-induced failures have not been an issue.

The transformation $t_{js}^{morph3D \to 2D}$ needs to accomplish the representation transformation between the kinematic model and the edge representation of the image discussed early in this paper. It therefore needs to map from each skeletal segment to the 3D surface defining that segment's canonical shape, generate one morph of that canonical shape and then project that morph onto the viewplane for the particular 3D rotation. There are two forms:

(a) Assumes rotation in azimuth and elevation is applied to the skeleton, as indicated in (34), for example. In this formulation $t_{js}^{morph3D \to 2D}$ has one parameter, which is *morph_variant*. Here, early in the convergence morphing is applied to superpositions of rotation hypotheses, and this imposes special requirements on the transformation (whose technical details are outside the scope of this discussion).

(b) Assumes the 3D rotation to accommodate viewpoint takes place in the same layer as the morphing and projection to 2D. Consequently each $t_{js}^{morph3D \to 2D}$ has, in this formulation, two implicit parameters: *morph_variant* and *view_direction*. In this case the morph is applied only to a single rotation and the superposition of viewpoints occurs in the 2D projection domain.

The set of $t_{js}^{morph3D \to 2D}$ transformations combine what would in theory could be two layers, but would require an inverse in the forward path, $t_{js}^{2D \to 3D}$, whose parameter is *viewing_direction*, which takes the edges of the 2D image to a 3D surface. The latter, of course, is indeterminate unless there is a correct surface onto which (actually tangent to which) the 2D edges can be projected. The latter is available if in the backward path there is a layer which takes each skeletal segment hypothesis and "fleshes it out" to the canonical surface and then morphs it, i.e. $t_{js,morph\_varient}^{canon+morph}$. The transformation in the latter layer has the single parameter *morph_variant*. The backward and forward equivalents are defined in (35).

$$\begin{aligned} t_{js}^{morph3D \to 2D} \circ T^{azim \times elev\,rot}(\mathbf{m}^s) &= t_{js,view\_dir}^{3D \to 2D} \circ t_{js,morph\_varient}^{canon+morph} \circ T^{azim \times elev\,rot}(\mathbf{m}^s) \\ t_{js}^{demorph2D \to 3D} \circ I^{2D\,rot} &= t_{js,morph\_varient}^{morph \to canonical} \circ t_{js,view\_dir}^{2D \to 3D} \circ I^{2D\,rot} \\ \text{where } I^{2D\,rot} &= T^{2D\,rot} \circ T^{scale} \circ T^{shift}(I) \end{aligned} \tag{35}$$

However, the practical implementation of $t_{j_s morph\_varient}^{morph \rightarrow canonical}$ involves technical difficulties, so in the implementation used for the demonstrations in this paper the form shown in (34) is used.

As discussed earlier, the configurations in (34) and (35) will not reproduce the human behavior of interpreting pose and motion from markers on the joints. This could imply that there is a parallel path which locates the axes of the limbs and possibly the neck and head in the 2D image and bypasses the morphed and canonical 3D representations but maps directly to the kinematic skeleton. Why, from an evolutionary point of view, this should exist is a bit of a mystery, but it may be that its relative simplicity recommended it as an earlier stage of evolution, to be later elaborated by the sequence of transformations to and from surface models. An alternative explanation is that the skeleton representation persists within the canonical surface representation so stick figure images and the minimalist version, joint marker images, can directly map through to the skeletal segments.

Figure 6 illustrates the effect of the morphing and articulation transformations on the canonical body model and their relationship to the input image.

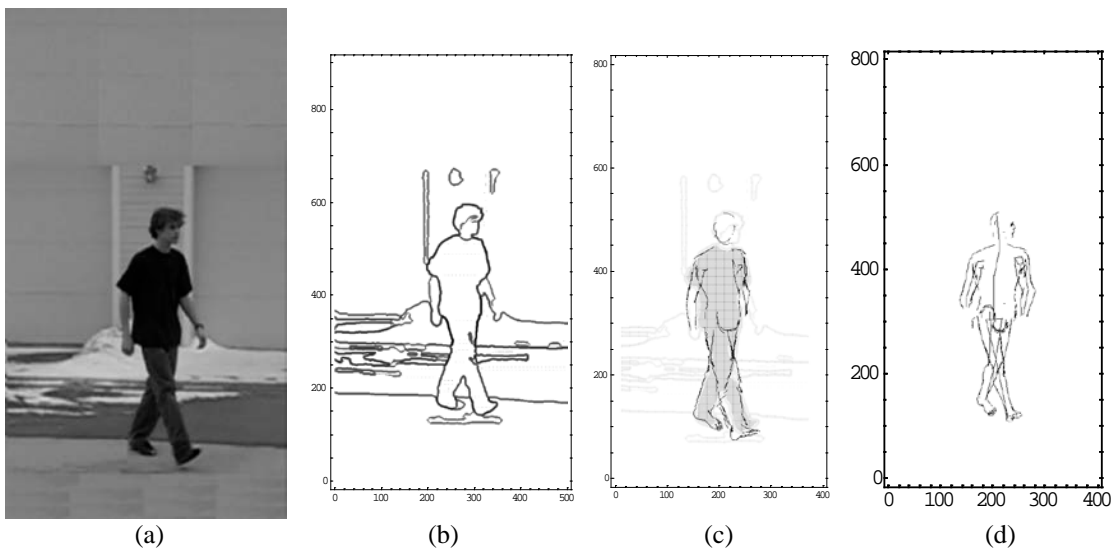

(a)            (b)            (c)            (d)

Figure 6. An example of the morphed and unmorphed solutions and their relationship to the figure in the input image. (a) input image, (b) edge filtered input image, (c) the morph of the solution pose of the canonical model, (d) the solution pose of the canonical model with the skeletal segments shown. In (c) the grayed area corresponds to the black clothing in the input image. The difference in the contours between (c) and (d) is the effect of the morphing transformation. The close congruence of the morph solution for both the clothing and the arms is apparent. The head of the canonical model is not subjected to morphing in this implementation so the head orientation solution is not optimal.

### MSC stages for kinematic transformations

For the kinematic MSCs which compute the pose for each body segment chain, $\mathbf{f}^0$ and $\mathbf{b}^{m+1}$ are the spatial position possibilities for each end of the kinematic chain, determined from the input image and/or the proximal chain endpoint. The stages in between apply the articulation transformations in each kinematic chain in (7-12).

To determine the articulation parameters $k_1, \cdots, k_S$ which reconstruct a pose in the input image, the compositional representation of the skeleton comes into play. For this, separate MSCs or circuits model the kinematic chains in Equations 7-12, and some of the input and output stages of these are chained together. The root position $\mathbf{r}$ designates the single voxel which provides the anchoring in 3D space. Recall that these segment generating transformations are applied as mappings in a voxel-like 3D "image" of the space in which the skeleton is posed. A given transformation $\lambda$ maps all non-zero voxels to voxels at the specified displacement, thus generating the reach of a body segment at a particular orientation from any hypothesized voxel location of its proximal joint to the voxel location of its distal joint.

The articulations of each segment are constrained in 3D space by the input image along the lines of view of each segment's distal joint, as suggested by Figure 3. This constraint is enforced using the mask $\boldsymbol{\mu}^L$ in (25,28) to zero the voxels in each $\mathbf{f}$ and $\mathbf{b}$ vector which do not lie along these lines of view. This removes those voxels from the possible solution. In practice, because of occluded joints, the voxels are only inhibited, not zeroed. And the lines of view are blurred with a Gaussian to allow for some error in estimated joint position. In (36) $\mathbf{p}_k$ represents a particular configuration of non-zero voxels which are the possible proximal joint locations for the first segment of the chain. Similarly, $\mathbf{z}$ represents the voxel candidates of the endpoint of the last segment, either determined by the next kinematic chain, defined by the inverse kinematic goal or defined by the line of view of the limb endpoint. The computations of $q$ by MSC for a three segment kinematic chain are

$$\begin{aligned} q_{k_1}^{s_1} &= c\left(\Lambda'^2 \circ \Lambda'^3(\mathbf{z}), \lambda_{k_1}^1(\mathbf{p}_1)\right) \\ q_{k_2}^{s_2} &= c\left(\Lambda'^3(\mathbf{z}), \lambda_{k_2}^2 \circ \Lambda^1(\mathbf{p}_2)\right) \\ q_{k_3}^{s_3} &= c\left(\mathbf{z}, \lambda_{k_3}^3 \circ \Lambda^2 \circ \Lambda^1(\mathbf{p}_3)\right) \end{aligned} \quad (36)$$

where $\Lambda^s$ designates the aggregate of weighted pose transformations designated by $\lambda$'s, as defined in (31). When $\mathbf{p}$ contains multiple loci, kinematic chains are created from each of the starting loci. Just as the starting loci for successive segments are the endpoints of the previous segment, the starting loci for one kinematic chain can be the endpoints from another, as in (11) and (12). The computation of kinematic chains using segment transformation is more fully discussed in [9].

An example of MSC inverse kinematics applied to a physical 6DOF robotic arm (an approximate analog of the human arm's articulations) in the presence and absence of obstacles is seen in Figure 7 Like the human arm, a robotic arm of this architecture is redundant, meaning it can reach the same endpoint in a range of configurations. MSC first converges to the set of redundant solutions and then further converges to one of these, essentially arbitrarily, on the basis of numerical "noise."

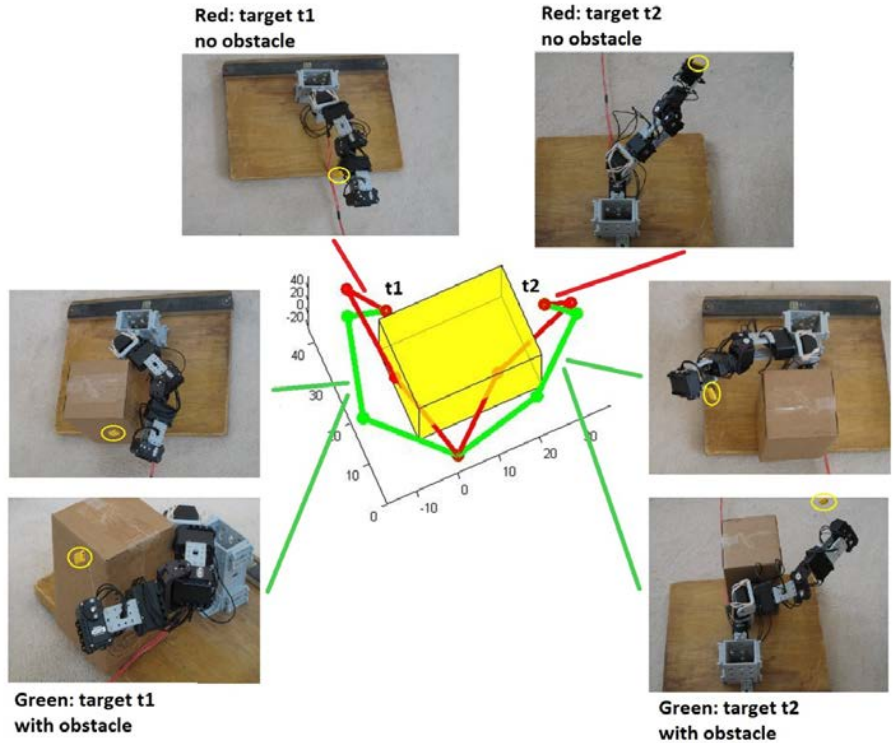

Figure 7. 6 Degree of Freedom robotic arm reach planning in the presence and absence of solid obstacles. Graphical depiction of MSC solution (center) for 6 DOF arm in presence of obstacle (green lines) and without obstacle (red lines) to two targets (coordinates for t1: -2, 35, 10 ; coordinates for t2: 30, 20, 25) . Images: Smart Servo 6 DOF arm, end effector is the circled yellow tag. Physical arm solutions are directly controlled by MSC solutions, as indicated. MSC solves the inverse kinematics in the shoulder reference frame. The resulting joint angles in shoulder space are converted to local joint angles (for controlling the servos) by simple vector computations.

**The System View**

The system for pose reconstruction/interpretation comprises circuits for visual, kinematic and morphology transformation discovery. Since each is an MSC, the circuitry of each differs only by the transformations implemented. As will be seen later, this difference at an anatomical level amounts to which synapses have strong weight and those which have zero weight or don't physically exist. Consequently, at the dataflow level of abstraction, all MSC's appear the same, except for the number of layers or stages. At the dataflow level of abstraction the system appears as presented in Figure 8, with the particular transformations for each layer indicated by the superscript on the **q** vector computed by each. The kinematic MSC is simplified to three layers, as if computing only one 6 DOF arm. In reality there are layers for each of the skeletal segments described earlier.

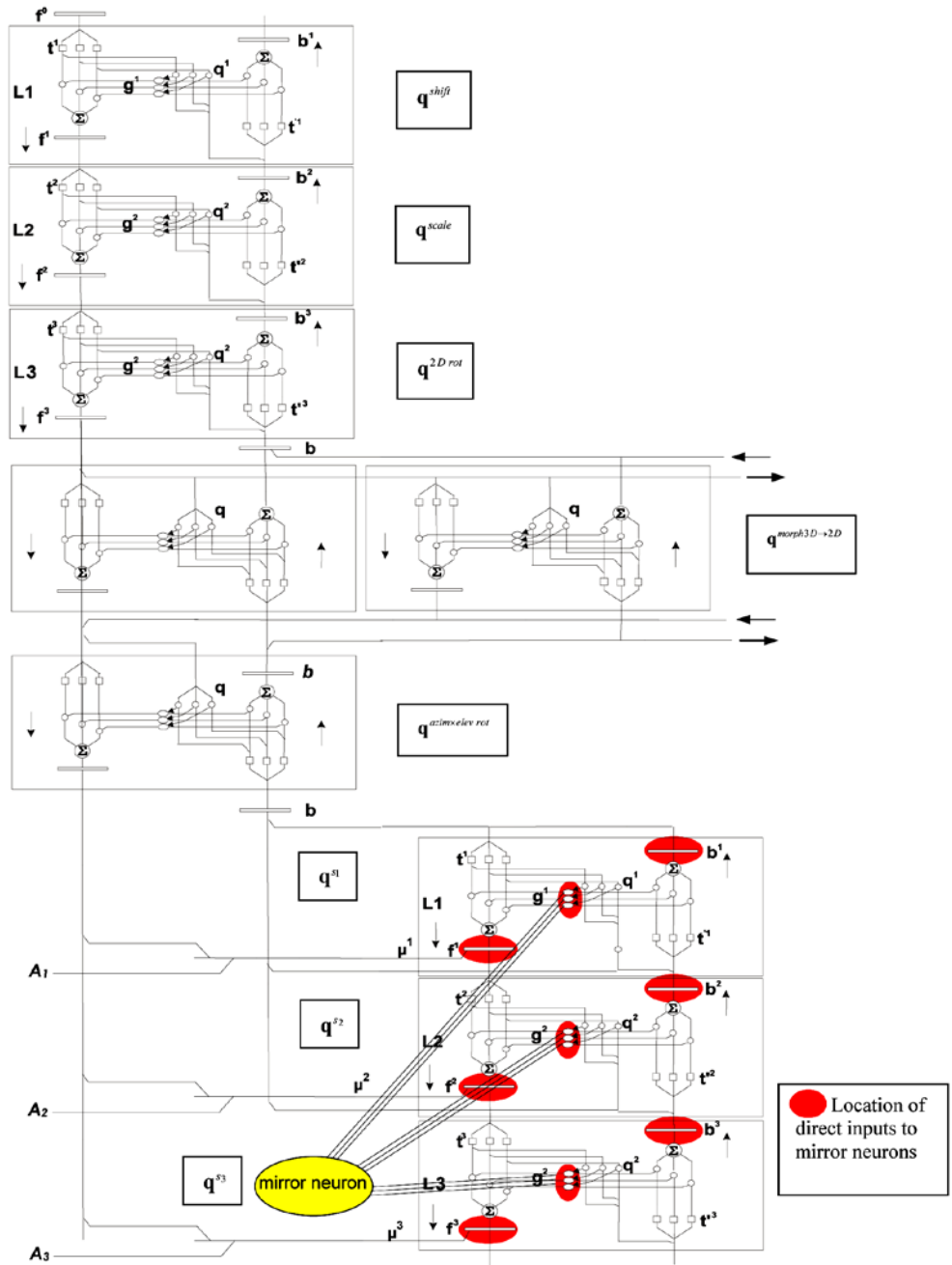

Figure 8. Combined visual, kinematic circuits with mirror neuron inputs. The diagram illustrates the dataflow of the computation. Each rectangle represents a stage of transformation. The pathways inside the rectangles indicate the dataflow for the computation of **q**, **g** and the gating of the transforms into superpositions in **f** and **b** vectors. Note, in visual circuit, forward is leftmost dataflow, backward is rightmost dataflow. The kinematic circuit receives input from either the visual circuit or the lines labeled $A_1$, $A_2$, $A_3$ which provide masking inputs for inverse kinematic targets. There are several possible variants for the layers that compute 3D rotation of the kinematic model and morph it to the full shape of the body in the input image. These are discussed in the text.

Circuitry exhibiting mirror-like behaviors would capture pose parameters, or for actions a time series of pose parameters, from **g** and **f** or **b** vectors from each kinematic circuit stage involved in the particular action. For mirror neurons exhibiting more complex conditional behavior, or capturing time sequences of poses, other inputs to the circuitry would be required as well.

**Reducing solution ambiguity with auxiliary constraints**

As pointed out earlier, for input imagery with no depth information from perspectivity or multiple viewpoints, the positioning of joints in depth can be ambiguous. Most ambiguities are excluded by the rotational geometry of the limbs and joint angle constraints because in theory there are only two points of intersection between the circular or spherical surface of a rotational 1 DOF or 2 DOF segment and the view line of end joint of that segment (assuming the root of that segment is fixed). But in practice the joint itself cannot be precisely located and the root of most segments themselves have multiple possible locii. Consequently, joint angle constraints are often not sufficient to eliminate the possibility of erroneous, even "absurd" reconstructions. Of course, a probabilistic approach could be taken, biasing solutions to the more common poses when there are several possible solutions. But that approach is rejected here on the principle that human observers are quite capable of interpreting rare, improbable poses and rarely if ever make mistakes in pose interpretation, even from poor quality photographs of distant figures. We can also almost always discern whether a pose is static or dynamic, stable or unstable. The primary clue for this discernment is the effect of gravity. That is, we can make the mental calculation from the assumed center of mass, whether the observed support for the pose allows the pose to be maintained without major adjustment. The simplest condition for static, standing poses is that the center of mass projected vertically onto the ground surface (whether plane or otherwise) lies along the line segment connecting the midpoints of the two feet. The position of the feet in the input image allows the line connecting them in 3-space to be computed, and the assumed locus of the root of the skeleton allows the center of mass in 3-space to be estimated. These allow the permissible end points regions for the supporting limbs on the ground surface to be delineated, and the IK solution by MSC is thereby constrained by masking (zeroing) all positions for the leg/foot endpoints except those permissible, as shown in Figure 9 for a plane ground surface.

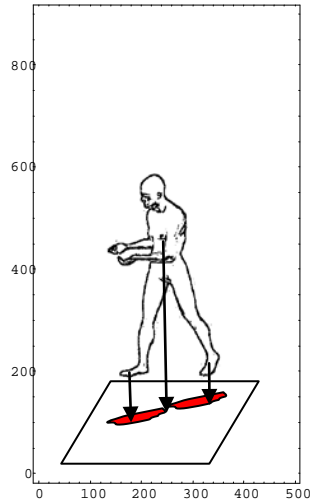

Figure 9. Projection of center of mass to ground plane, and estimation of permissible leg/foot endpoints (red).

The effect of applying this gravity constraint can be large or subtle, as shown in Figure 10. Note that the effect is only visible in viewpoints away from that of the original input image.

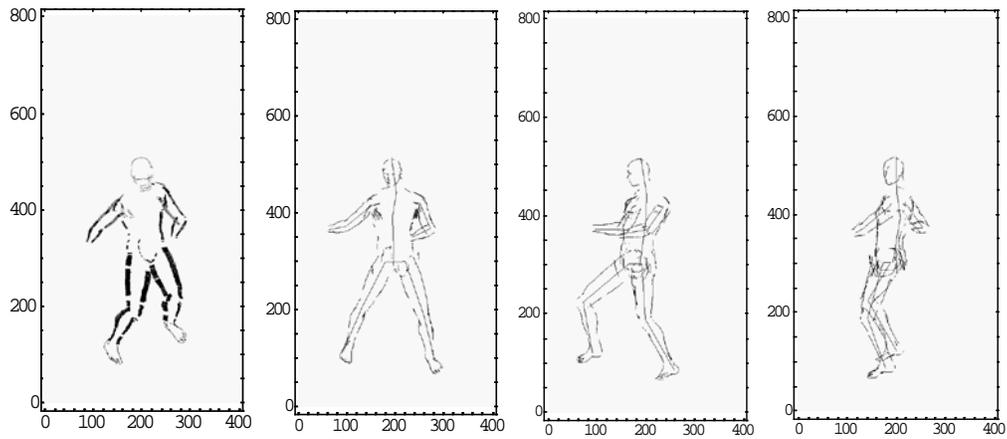

**(a) Gravity Constraint Disabled**

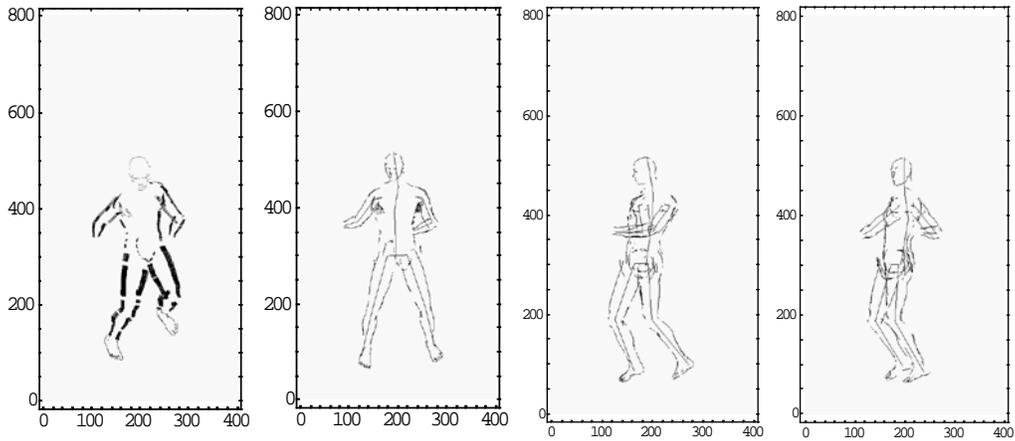

**(b) Gravity Constraint Enabled**

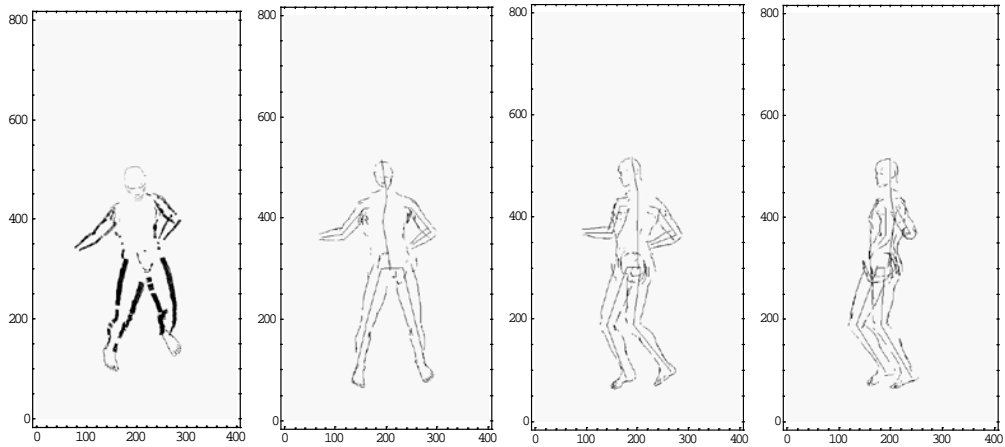

**(c) Gravity Constraint Enabled (Weight on balls of feet instead of heels)**

Figure 10. Effects of gravity constraint, or its absence, on 3D pose reconstruction. Leftmost panel in each row is the 3D reconstruction seen from the same viewpoint as the original input image. The remaining three panels of each row show three other viewpoints of the same reconstruction. (a) Pose reconstruction with gravity constraint disabled. (b) Pose reconstruction with gravity constraint enabled along line segment between heels. (c) Pose reconstruction with gravity constraint enabled along line segment between balls of feet. The awkwardness of the constraint-disabled solution is evident in the third and fourth panel, *though invisible from the original image viewpoint*. The subtle influence of the difference between the constraint-enabled weight on heels and weight on balls of feet solutions can be seen in the positioning of the upper body.

Of course for dynamic poses, this simple calculation does not suffice. For a runner banking on a curved path, the projection of the center of mass to the ground plane is tilted, not vertical, and typically lies to the medial side of the supporting foot. For other dynamic actions, gravity may not signify at all. And obviously for arm poses, center of mass does not provide a significant constraint. A full discussion of contextual pose constraints is outside the scope of this paper, but the principle of how such constraints are applied to restrict the solution space in an MSC implementation is the same as that described for gravity in static poses.

## A neuronal circuit isomorph of the MSC algorithm

The MSC algorithm has an almost one-for-one neuron-like circuit analog [9]. Each of the operations in the algorithm corresponds to a neuronal subcircuit which executes an equivalent operation. As in the algorithmic form, in the neuronal circuit mappings simply move the signal from a cell representing one locus to a cell representing another locus. As in the algorithmic form, the neuronal circuit uses a signal indicating the relative contribution of each mapping to the correspondence between forward and backward paths to gradually prune away the less useful mappings and preserve the most useful. Like the algorithm, the circuit is formed from a sequence of generic bidirectional stages which are differentiated only by the particular mappings they implement. Within each stage there is a neuron which implements each mapping and applies a gain coefficient $g$ to it, and a neuron which computes the equivalent of the $q$ associated with each mapping. There is a circuit which implements the updating of each gain coefficient by the corresponding $q$ and at the same time carries on the competition between $g$'s. For simplicity, the mask $\boldsymbol{\mu}$ is ignored in this discussion.

Though the neuronal circuit could in principle operate iteratively like the algorithm, it more naturally executes as a pipeline (as could the algorithm with suitable digital hardware) with all stages in both directions executing concurrently and passing results on to the next stage. The architecture therefore naturally leads to pulsed dynamics, but because of both the duration of the pulse and the concurrency of execution, each cycle of the analog circuit advances convergence considerably more than one iteration of the algorithm [9].

**Mapping or transformation**
The neuronal circuit bears some resemblance to a crossbar switch, as evident in Figure 11. A crossbar switch has many inputs and many outputs, and can at the same time connect each input to any single output. That connection mapping can be represented by a permutation matrix where each row corresponds to an output and is all zeros except for the column which specifies the input. While a crossbar switch implements every possible interconnection of inputs and outputs, the neuronal MSC only implements the connectivity necessary to implement the mappings. Consider only the forward path (blue in Figure 11). A group of $n$ neurons represents the forward vector for stage $L$-1, $\mathbf{f}^{L-1}$. Each of these neurons represents a pixel-like or voxel-like input to stage $L$-1. Their axons fully interconnect to the dendrites of a group of $m$ neurons representing the forward vector for stage $L$, $\mathbf{f}^L$. That is, there is an excitatory synapse between each axon of $\mathbf{f}^{L-1}$ and each dendrite of $\mathbf{f}^L$. However, these synapses are such that they do not produce a post-synaptic potential unless a concurrent activation is present on a paired excitatory gating synapse on $\mathbf{f}^L$. The paired synapses are represented as opposing arrowheads in Figure 11.

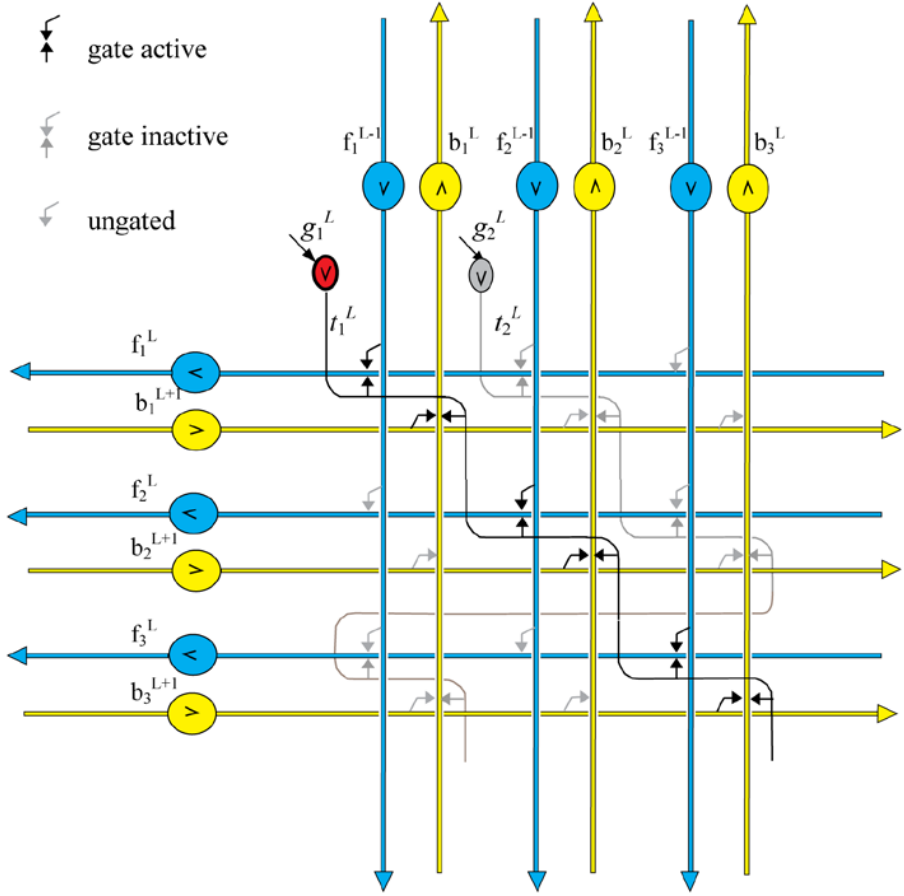

(a)

$$\mathbf{f}^L = \mathbf{T}_1^L \mathbf{f}^{L-1} \quad \mathbf{T}_1^L = \begin{vmatrix} 1 & 0 & 0 \\ 0 & 1 & 0 \\ 0 & 0 & 1 \end{vmatrix} \quad \mathbf{f}^L = \mathbf{T}_2^L \mathbf{f}^{L-1} \quad \mathbf{T}_2^L = \begin{vmatrix} 0 & 0 & 1 \\ 1 & 0 & 0 \\ 0 & 1 & 0 \end{vmatrix} \quad \mathbf{b}^L = \mathbf{T}_1'^L \mathbf{b}^{L+1} \quad \mathbf{T}_1'^L = \begin{vmatrix} 1 & 0 & 0 \\ 0 & 1 & 0 \\ 0 & 0 & 1 \end{vmatrix} \quad \mathbf{b}^L = \mathbf{T}_2'^L \mathbf{b}^{L+1} \quad \mathbf{T}_2'^L = \begin{vmatrix} 0 & 1 & 0 \\ 0 & 0 & 1 \\ 1 & 0 & 0 \end{vmatrix}$$

(b)

Figure 11. Neuronal mapping circuit for a single stage: (a) forward path (blue) and backward path yellow) with transformations implemented by mapping neurons, (b) matrix-vector equivalent operations. Neuron body is represented as an ellipse with direction of signal flow indicated by the enclosed ∧. Blue or yellow lines with arrowheads are axons, those without are dendrites. The axons of mapping neurons with inputs labeled $g_1^L$, $g_2^L$ implement transformations $t_1^L$, $t_2^L$ by gating selectively between $\mathbf{f}^{L-1}$ and $\mathbf{f}^L$, and between $\mathbf{b}^{L+1}$ and $\mathbf{b}^L$ as described in the text. Arrows at intersections of like-colored axons and dendrites indicate excitatory synapses. Signals are only transferred when the synapses between axon and dendrite of the same direction (yellow or blue) also receive an active gating signal from the axon of a mapping neuron. By enabling corresponding synapses in both forward and backward paths, each mapping neuron enables both a forward and inverse mapping.

Each forward and backward (inverse) transformation pair, $t_j^L$ and $t_j'^L$, in stage $L$ is implemented by a single mapping neuron. Each mapping neuron activates many gating synapses: up to one on each dendrite of $\mathbf{f}^L$ (and, as will be seen shortly, up to one on each dendrite of $\mathbf{b}^L$). The signal from one mapping neuron causes the necessary rearrangement of signals from the $\mathbf{f}^{L-1}$ neurons to be gated onto the $\mathbf{f}^L$ neurons. The connections gated by the mapping neuron for $t_j^L$ can be represented by the non-zero elements of the permutation matrix $\mathbf{T}_i^L$. The transform implemented by mapping neuron for $t_j^L$ is $\mathbf{T}_i^L \mathbf{f}^{L-1}$. (Note, the matrix notation for a single transformation $\mathbf{T}_i^L$ should not be confused with the algorithmic superposition notation $T^L$.)

The backward path (yellow in Figure 11) is completely reciprocal to the forward path. A group of $m$ neurons represents the backward vector for stage $L+1$, $\mathbf{b}^{L+1}$. The axons of $\mathbf{b}^{L+1}$ are fully interconnected with a group of $n$ neurons which represent the backward vector for stage $L$, $\mathbf{b}^L$. As in the forward path these synapses do not produce a post-synaptic potential unless a concurrent activation is present on the paired gating synapse on the $\mathbf{b}^L$ dendrite.

Each forward path transformation $t_i^L$ is paired with a backward path transformation $t_i'^L$ which "undoes" the former. Mathematically, instead of transposing $\mathbf{T}_i^L$ to obtain this inverse mapping, imagine redefining matrix multiplication by exchanging the row and column indices of each operation. Then, using the new rules, $\mathbf{T}_i^L$ can be used to compute $t_j'^L(\mathbf{b}^{L+1})$. This is what the circuit does on the backward path. In the circuit this is conveniently implemented by activating the gating synapses on the $\mathbf{b}^L$ dendrites with the same "coordinates" in the interconnect mesh as the gating synapses which implement the associated forward path transformation. Due to the proximity of the forward and backward gating synapses, this can be readily done by the same mapping neuron if it has enough synapses. (If one neuron does not have enough synapses, several mapping neurons can be combined to implement a transformation pair.)

Though the $\mathbf{f}^{L-1}$ to $\mathbf{f}^L$ and $\mathbf{b}^{L+1}$ to $\mathbf{b}^L$ connectivity in Figure 11 is illustrated as all-to-all, in fact the necessary connectivity is more sparse: there need be connection only at those intersections where a mapping neuron synapses, that is, where at least one matrix $\mathbf{T}_i^L$ in the stage has a non-zero entry.

**Gating**

Now, consider that the non-zero elements of the permutation matrix $\mathbf{T}_i^L$ are all equal to $g_j^L$ instead of 1.0. On the forward path $\mathbf{T}_i^L \mathbf{f}^{L-1}$ computes $g_j^L t_j^L(\mathbf{f}^{L-1})$ and the index-exchanged matrix-vector multiply on the backward path computes $g_j^L t_j'^L(\mathbf{b}^{L+1})$. To implement this, the activation of the mapping neuron must have an amplitude such that the post-synaptic signal is proportional to $g_j^L$. It does not matter if the actual gating mechanism is not a perfect multiplication so long as it is monotonic (in the sense that if at least one argument of the operation increases and neither decreases, the result increases).

No additional circuitry in Figure 11 is needed to implement the "gain control" of each transform signal so that it is proportional to its *g* coefficient.

**Aggregation**

Suppose for simplicity that the dendrite of each $\mathbf{f}^L$ and each $\mathbf{b}^L$ neuron strictly adds all its inputs. Now, when more than one mapping neuron is active the signal from the dendrites of $\mathbf{f}^L$ and $\mathbf{b}^L$ are the sums of the weighted transforms.

$$\mathbf{f}^L = g_1^L \cdot t_1^L\left(\mathbf{f}^{L-1}\right) + \cdots + g_n^L \cdot t_n^L\left(\mathbf{f}^{L-1}\right) \text{ and } \mathbf{b}^L = g_1^L \cdot t_1'^L\left(\mathbf{b}^{L+1}\right) + \cdots + g_n^L \cdot t_n'^L\left(\mathbf{b}^{L+1}\right) \quad (37)$$

In the notation used earlier the *g*'s and all transformations were captured in one symbol, *T*, so this would read

$$\mathbf{f}^L = T^L\left(\mathbf{f}^{L-1}\right) = \sum_i g_i^L \cdot t_i^L\left(\mathbf{f}^{L-1}\right) \text{ and } \mathbf{b}^L = T'^L\left(\mathbf{b}^{L-1}\right) = \sum_i g_i^L \cdot t_i'^L\left(\mathbf{b}^{L-1}\right) \quad (38)$$

Therefore the simple circuit shown in Figure 11 implements not only mapping and gating, but aggregation as well. More realistically, the signal strength in a given $\mathbf{f}^L$ or $\mathbf{b}^L$ neuron dendrite is not a strict superposition, but a more complicated sum-like *aggregate* of its gated inputs, and the axonal output is proportional to that. The aggregation function denoted $\alpha$ should ideally reflect the dendritic arborization and dynamic properties and the transfer function to the axonal signal

$$\begin{aligned} \text{forward:} \quad & \mathbf{f}^L = \alpha\left(g_1^L \cdot t_1^L\left(t_1^L \mathbf{f}^{L-1}\right), \cdots, g_n^L \cdot t_n^L\left(\mathbf{f}^{L-1}\right)\right) \\ \text{backward:} \quad & \mathbf{b}^L = \alpha\left(g_1^L \cdot t_1'^L\left(\mathbf{b}^{L+1}\right), \cdots, g_n^L \cdot t_n'^L\left(\mathbf{b}^{L+1}\right)\right) \end{aligned} \quad (39)$$

**Correspondence**

In simplified terms, the neuronal correspondence-computing circuitry seen in Figure 12 implements something like a dot product.

$$q_i^L = c\left(\mathbf{b}^{L+1}, t_i^L\left(\mathbf{f}^{L-1}\right)\right) \approx \mathbf{b}^{L+1} \bullet t_i^L\left(\mathbf{f}^{L-1}\right) \quad (40)$$

There is correspondence computing neuron for each mapping neuron. Paired synapses along each mapping neuron dendrite (also represented by opposing arrows in Figure 12) compute the correspondence multiplications. One synapse of each pair receives input from an axon of $\mathbf{f}^{L-1}$ and the other from an axon of $\mathbf{b}^{L+1}$. The transformation $t_i^L$ is implemented by having the correspondence neuron synapses follow the same pattern in the interconnect mesh as the associated mapping neuron. The postsynaptic signals from each synapse pair are aggregated in the dendrite. Figure 13 shows the mapping and correspondence circuits combined, as they would occur in a single stage.

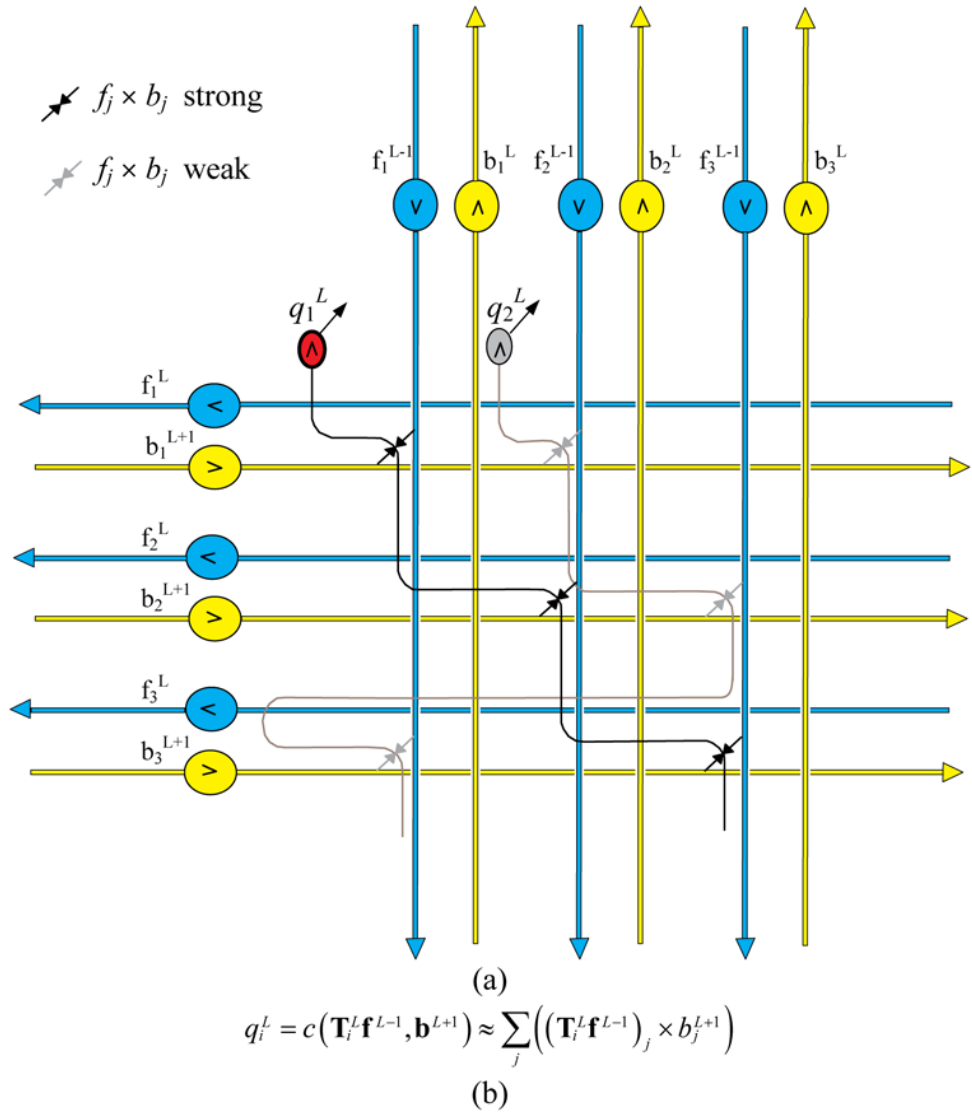

(a)

$$q_i^L = c\left(\mathbf{T}_i^L \mathbf{f}^{L-1}, \mathbf{b}^{L+1}\right) \approx \sum_j \left(\left(\mathbf{T}_i^L \mathbf{f}^{L-1}\right)_j \times b_j^{L+1}\right)$$

(b)

Figure 12. Neuronal correspondence circuit: (a) correspondence ($q$) computing neurons and (b) the matrix-vector notation for the operation. Forward and backward path neurons are the same as in Figure 11. Neurons labeled $q_1^L$, $q_2^L$ implement the computation of the correspondences of signals on axons of $\mathbf{f}^{L-1}$ and $\mathbf{b}^{L+1}$. The synapses between those axons and the dendrites of the correspondence computing neurons are located at the same intersections as the synapses of the associated mapping neurons, as illustrated in Figure 13. A correspondence neuron therefore implements the same transformation as its associated mapping neuron.

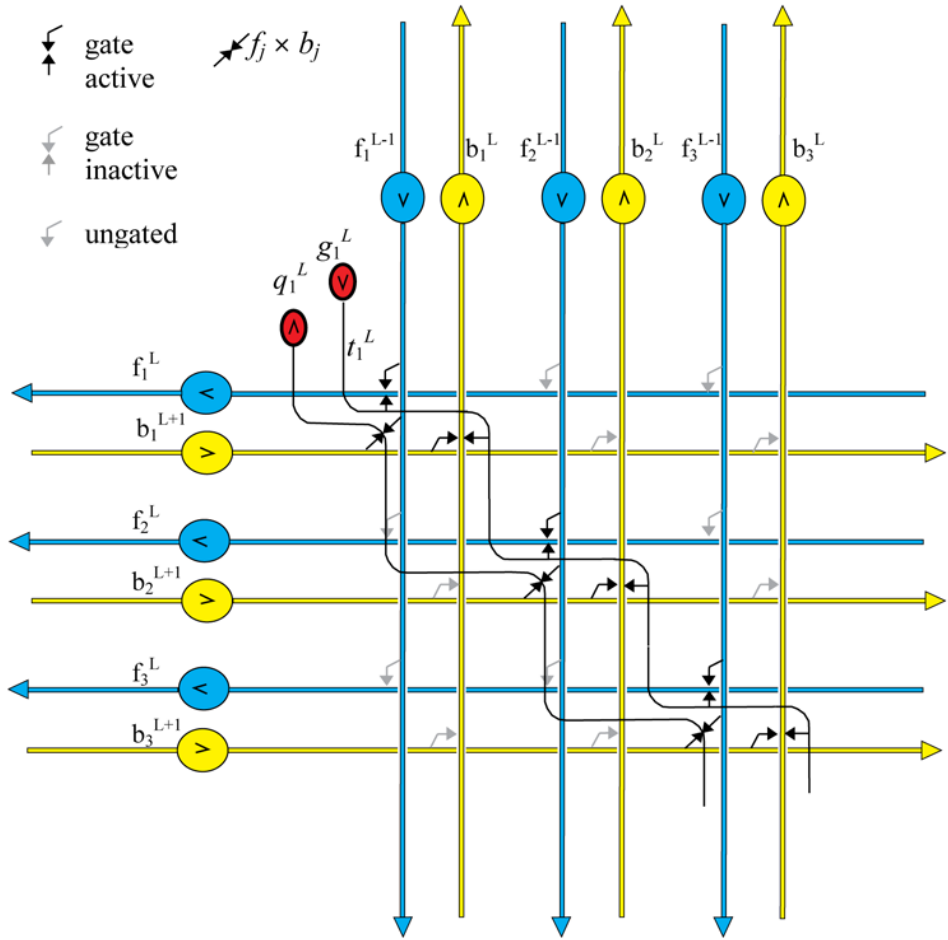

Figure 13. Mapping and correspondence circuits combined (only one transformation shown). See legends for Figures 11 and 12.

**Temporal aspects of circuit signals**
Before moving on to the competition subcircuit it is necessary to explain how signals are encoded in the neuronal circuit as a whole. While the circuit could be made to work with any encoding from DC to spike rate, phase coding is the simplest. That is, when the **f** or **b** neurons spike, it is assumed that the spikes for the neuron that represents the largest value will spike first, with lesser values lagging behind it monotonically in order of decreasing value. If a spike passes through a weaker synapse, or a synapse with weaker gating from a paired synapse, it will be retarded with respect to a spike of the same original latency passing through a stronger synapse or one with stronger gating. Thus, if one were to record from all the **f** neurons in a stage simultaneously, one would see small differences in the rise time initially. But as the gain coefficient on some decreases, those signals will fall increasingly behind the leading signals. If we introduce an inhibitory signal which is launched with the first arriving spike and then increases as more spikes arrive, that inhibitory signal can be used to increasingly retard and ultimately eliminate the later arriving signals. This is how the competition between mappings takes place, as will be described in more detail. The net effect is that in initial cycles, or iterations, many signals are active, but quickly the competition eliminates all but the ones that correspond to the best mapping. The signal dynamics over several cycles is seen in Figure 14, with the winning signal circled where its lead becomes visible. Since the interactions that implement this competition take place inside the dendrites, the waveforms that appear in Figure 14 are roughly characteristic of dendritic signals.

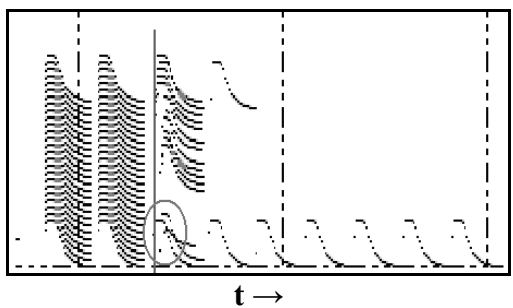

t →

Figure 14. Dynamics of temporal or phase coding under competition. Signals shown are typical of **f** and **b** vector elements. Each trace is one element of **f** or **b** vector. A trace on the graph disappears when its signal goes to zero. Hence the traces for many elements of the vectors disappear as convergence proceeds and some *g* coefficients become zero. Small differences in phase are amplified by competition until inhibition can eliminate all but the leading signal. The first visible phase lead of the eventual winner is marked with the oval. (Image greatly magnified to show temporal lags.) Figure from [9] is generated by a large scale, low resolution simulation of a neuronal MSC recognizing one of several 2D figures in an input image.

**Competition**
All that remains to complete the execution of the algorithm is the circuitry to compute the updates to *g* from (29) and (30).

$$\Delta g_j^L = -k \left( 1 - \frac{q_j^L}{\max(q_1^L, \cdots q_{n_L}^L)} \right)$$

$$g_j^L \leftarrow \max(g_j^L + \Delta g_j^L, 0)$$

The purpose of the expression $1 - q_i^L / \max(q_1^L, \cdots q_{n_L}^L)$ is to generate a signal which is minimal when $q_i^L$ is the largest among the *q*'s in stage *L*, and increases as $q_i^L$ decreases from the maximal value among the *q*'s. In the competition circuit in Figure 15, the first arriving signal on $\mathbf{b}^{L+1}$ sets off an increasing signal *p* which increasingly inhibits each subsequent $q_i^L$ the later it arrives: *p* is near zero for the first arriving *q* and increasingly large for the *q*'s that arrive with greater latency. *p* is applied as an inhibitor of the correspondence neuron input, *q*, to each mapping neuron, *g*. The mapping neurons, *g*, are different from most of the other signals in the circuit in that they hold a gradually changing value over several cycles. This longer time constant can be seen clearly in the dynamics in Figure 16. The signals $g_i^L$ and $g_j^L$ both start at maximum, but over the course of several cycles diverge to a winning and losing state, respectively. As a result the signals on $\mathbf{f}^L$, which start at approximately the same levels, become segregated into winners and losers.

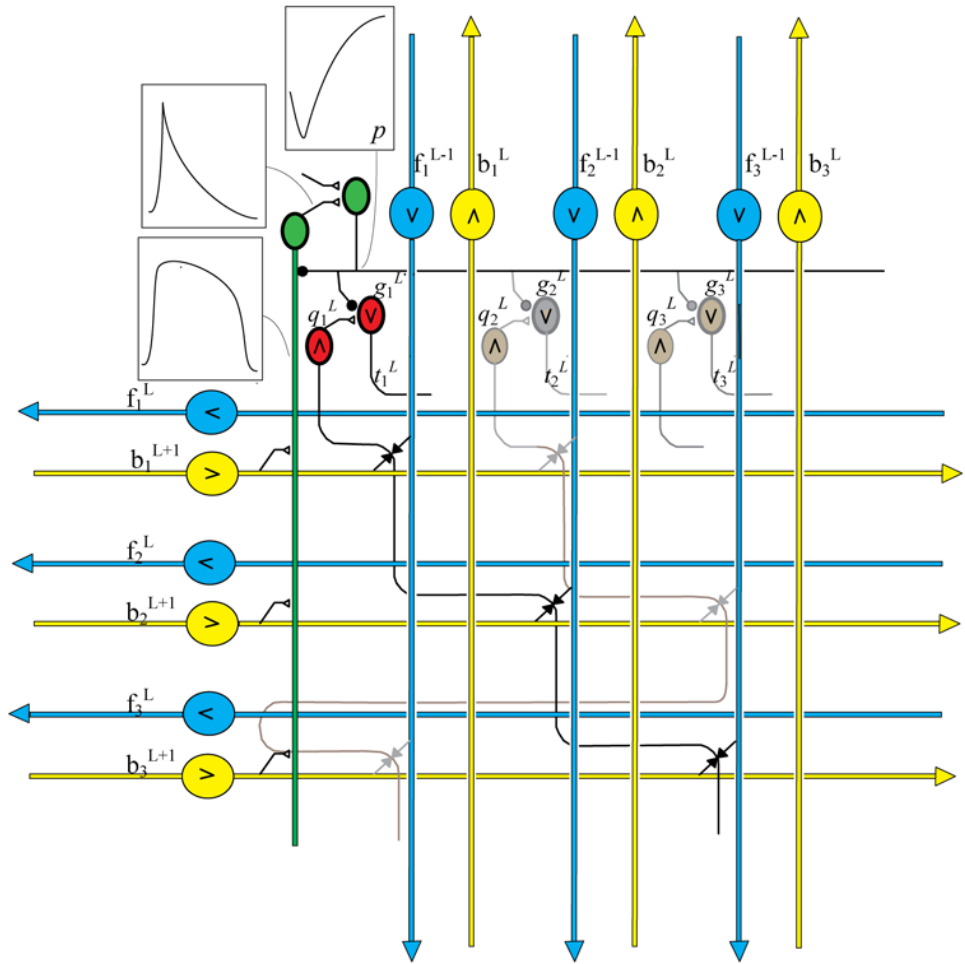

Figure 15. Neuronal competition circuit. The green neurons implement the dynamics which gradually increase the inhibition of the mapping neurons, $g_i^L$, thereby delaying the effective gating of the transforms they implement. The degree of delay is inversely related to the excitation by the associated correspondence computing neurons, $q_i^L$. The signal levels over the period of each cycle are indicated in the boxes. The effects of this circuit over several cycles are seen in Figures 14 and 16.

The oscillatory characteristic seen in Figures 14 and 16 may either arise from an external "clocking" signal or from forward-backward loop dynamics in each stage. This also implies that either the **f** or **b** neurons (or both) have a short duration sample-and-hold behavior, such that the signals they emit in each cycle are proportional to the excitation they have received in the previous cycle. There are a variety of ways this behavior might arise, but that discussion is beyond the scope of this paper.

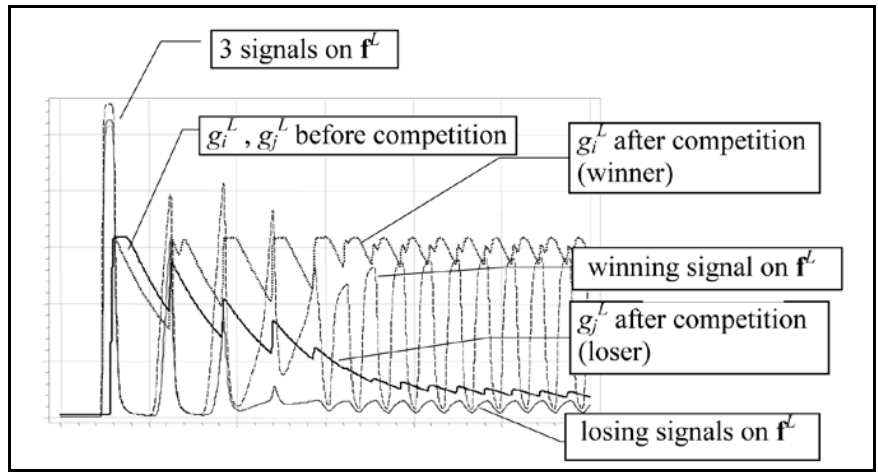

Figure 16. Neuronal circuit dynamics (high resolution Spice simulation of neuronal MSC). The winning and losing transforms are resolved by large differences in amplitude after several cycles. Note that no thresholds are applied in this simulation, so the signals of the losers survive at low amplitudes. If thresholds are applied the losing signals have zero amplitude in the steady state, and the steady state is reached earlier. Thresholds set too high risk incorrect convergence due to collusion, as discussed in the text. Figure from [9].

**Estimating neuron population and latency from the circuitry**
The uniformity of the MSC portions of the complete circuit make it fairly easy to estimate the neuron populations necessary to implement the various stages. Assuming the roughly 600 thousand transformations enumerated earlier, some 1.2 million neurons are required for mapping and correspondence. The **f** and **b** vectors for the 2D visual stages require about 300 thousand for $f^0$ and some fifty thousand each after $f^0$. The total for the visual circuit is about 2 million. This estimate assumes about 600 "pixel" diameters for $f^0$ and 150 for subsequent stages, and four orientation "channels" depth. The total probably needs to be multiplied by 5 to account for axon synapse counts, and 4 channels is almost certainly an underestimation. So the realistic count is probably in the 10s of millions. But this circuit would be shared by many visual functions beyond pose reconstruction and only accounts for a fraction of visual area neurons.

The kinematic space is harder to estimate because it depends greatly on spatial resolution. In practice this would be non-uniform, because resolution need for upper body and leg movements is obviously less than required for arms and hands. Uniform spatial representation is not required in either algorithmic or neuronal MSC because the mappings can accommodate metric variation. A coarse discretization of the "personal space" of about 300Hx200Wx100D suffices for full body pose solutions, though no stage requires more than part of that space. Some 10 million or more neurons would be required. Subsequent local refinement of the discretization would be required for finer motor tasks. Alternatively, each kinematic chain could be solved in a separate, smaller space, and then translated at far coarser resolution into a common space for collision detection. To this must be added storage for the morphing templates. In theory, this part of the model could engage some 15 million neurons, or fewer if partitioned by kinematic chain. This estimate ignores biologically realistic redundancy which would multiply the sum by a coefficient determined by the neurons' probability of generating an axonal signal when provided sufficient dendritic input.

An estimate of latency can be inferred from the circuit. In Figure 14 the circuit is seen converging to a single solution in five cycles. This timing is for a two or three stage circuit. Each extra stage adds a cycle. For the visual MSC this implies convergence in about seven or eight cycles. For the kinematic MSC, depending on how long a chain of segments needs to be solved, it can take from three to seven cycles. If one borrows 40Hz, or 25msec/cycle, from neurophysiology, one can estimate convergence in 200msec or less.

**Simulation**
The neuronal simulations which produced Figures 14 and 16 incorporated phase coding and highly non-linear limited dynamic range elements. Neither incorporated propagation delays. The largest scale simulation was limited by computational resources to a complete stage and two partial stages. The partial stages computed gain weighted superpositions from multiple templates and correspondences. The multiple templates acted as either multiple memories or multiple transforms. Thus the behavior being simulated was essentially a piecewise simulation of a whole functional circuit, but was complete from the point of view of simulating convergence dynamics. The lack of realistic neuronal propagation dynamics was the major shortcoming, but this is a common problem in large scale neuronal simulation. Nevertheless, full circuit latencies can be roughly inferred from the circuit architecture, as done above.

The most important property demonstrated by the neuronal simulations is that the convergence dynamics and ability to reach a correct solution tracked the algorithmic MSC despite the use of phase coding and non-linear, saturating circuit elements (whether simulated neurons in the low-resolution simulation or CMOS transistors in the Spice simulation). The reason the algorithm and circuit are so tolerant of variations in mathematical specifics is that the behavior is governed by ordering rather than precise quantitative relationships. That is to say, given three quantities, $a$, $b$ and $c$, an algorithm governed by ordering only requires that the relationship $a > b > c$ be preserved to obtain a correct solution. This is a much less demanding requirement than requiring that the differences $a - b$ and $b - c$ be preserved with high precision.

**Relationship to Prior Work**
While the author is unaware of any other computational neuroscience modeling efforts on pose or action reconstruction, the work reported here is related to a large body of work in machine vision on this task. We exclude the obvious Microsoft Kinect and similar pose extraction algorithms because the input to those is a 3D image instead of a 2D image, and this dramatically simplifies the problem. For systems using 2D image input there have been two classes of approach to the problem: model-based and appearance-based. The model-based approach relies on a linked segment model as a constraint on the interpretation of the shapes in the image. The approach here falls in this category. Appearance-based methods generally associate the body shape outline with a known skeletal pose. We do not consider the numerous approaches which require hand marking of the joint locations to be comparable, as this is precisely one of the great difficulties of the problem. Unlike the method described here, most unassisted model-based approaches make an initial pass over the image to segment candidate body part shapes, using such methods as superpixelization [20], Delaunay triangulation [21], or shape-contexts[22]. Due to the difficulty of the problem of pre-segmenting, several use what could be considered semi-manual parts marking: simulated part detectors of specified probability [23] or pre-specifying where skin patches and/or known clothing boundaries are expected [24]. Then the skeletal model is fitted to these segments by adjusting skeleton geometry parameters, usually while satisfying hard connectivity constraints, but in one case soft "attraction" [23]. Because of the dimensionality of the problem, the skeletal model is often 2D [20, 21], deferring the problem of "lifting" the 2D solution to the third dimension to a subsequent step [22]. The disadvantage in fitting a 2D model initially is that the natural joint-angle constraints and other 3D dependent constraints such as self-interference and gravity cannot be applied to help localize the joint locations. Exploiting these constraints is a major advantage of fitting the 3D model directly, as here and [23, 24]. The method presented here differs from most other model-based approaches known to the author in that the segmentation of the image and model parameter estimation emerge concurrently. MSC's ability to tractably search the full visual transformation space concurrently with the kinematic transformation space makes this possible. The adaptive approach to morphing used here is most similar to that used by [25]. All the model-based approaches, including the method presented here, devote significant computational effort to the problem of mapping the image body shape to the skeletal model, while at the same time mitigating the ambiguity that degree of freedom creates. Most use some consistency and constraint evaluation, both on the pose and on the properties of the image patch surfaces to reduce the ambiguity. A review of earlier work in this area can be found in [26].

Much of the model-based literature is devoted to interpreting pose and action from video image sequences or multiple cameras [27]. The problem of determining initial pose, as mentioned in the introduction of this paper, is well recognized. In some cases the initial pose must be hand-initialized [28]. Others disambiguate the initial pose estimate by reconciling it with poses in subsequent frames, given the assumption of movement continuity [29], or cleverly, from sharply cast shadows [30]. From a psychophysical perspective, these techniques seem less relevant than those mentioned earlier, given human ability to accurately interpret pose from a single image, with or without shadows or shading.

The appearance-based techniques use a variety of feature filters or shape descriptors to characterize a library of poses or actions of isolated figures, e.g. [31, 32]. The encoding of each image or sequence is associated with a set of 3D pose parameters. The same filters or descriptors are then used to parse the objects in the input image and the resulting representations are used to retrieve the pose parameters by various means of association. Most models in this category are proposed as machine vision techniques, but several recent models are biologically inspired feedforward models with spatio-temporal filters to detect motion signatures [33, 34]. These are implemented as classifiers, but in principle the classification could respond with kinematic parameters if such were associated with the training set. However, this leaves unanswered the question of how a biological system would acquire those parameters during visual training, since appearance-based methods cannot make *de novo* kinematic interpretations. This points up a contradiction between an essential property of mirror neurons and appearance-based models. Since the latter have no mechanism for viewpoint transformation, they provide no explanation for how the mirror neuron, or the system which provides its inputs, learns to associate an observed action with its own motor representation of an action, since the animal cannot observe its own action from the viewpoints it observes the action executed by others.

The system proposed here connects well with a number of the earlier efforts at mirror neuron modeling. Some of these incorporate cameras and physical robotics [35, 36, 37]. A review of that literature as of 2006 appears in [38]. The work here presents an alternative (and perhaps more biologically plausible) method to computing the inverse and forward kinematics used in a number of those models, e.g. [39]. While the inherent problem of visual transformation is well recognized, the models reviewed in [38] and some of the more recent developments of those models, either ignore the issue by assuming a constant viewpoint [35] or by precomputing self- and other- viewpoint transformations [36, 37]. The approach taken here to the vision part of the problem alone would extend such kinematics-based mirror neuron models in a natural way.

The work presented here brings several new aspects to the model-based category. First, from a mathematical standpoint, MSC's additive scaling property dramatically mitigates the combinatorial issues inherent in the dimensionality of the model parameter space. Hence an MSC-based model can tackle the problem of accommodating the full range visual transformations, which has been somewhat ignored in vision neuroscience models, as pointed out in [39]. This scaling property both eliminates the need to pre-segment the image and allows a direct search of the full 3D kinematic range of motion. As a biological model its concurrent use of top-down and bottom-up inputs maps well to the known anatomy, and its top-down driven attentional behavior is inherent [40]. Also, given the anatomical similarity of cortical circuits across the brain, the application of a single algorithm to both the visual and kinematic parts of the problem is attractive. In fairness, most machine vision model-based approaches make no claim to biological relevance, so comparison with the MSC-based technique should consider only criteria relevant to machine vision. However, the existence of a neuronal circuit analog of MSC leads to neuroanatomical, neurophysiological and psychophysical predictions comparable in specificity to those of the bio-inspired appearance-based models. These predictions are discussed in the next section.

## Implications

The hypothesis presented here of course makes a specific prediction about the wiring of both visual and kinematic circuits, and the similarity of both, though it may be a long time before predictions of this specificity can be tested. The most accessible direct consequence of the hypothesis is that many neurons in the motor planning cortices should be engaged in interpretation of observed pose or action, and that many of these should exhibit activity across a wide variety of poses or actions. Another accessible consequence is that during pose reconstruction, convergence dynamics in both visual and motor cortices should look quite similar and unfold over similar time scales, starting with a burst of densely distributed activity and quickly sparsifying over the latency interval of interpretation (Figure 17f). Due to the differences between visual and kinematic mappings the activity might be denser in the visual cortices involved. On the other hand, for actions or poses that require solutions for long kinematic chains, e.g. from foot to hand, the convergence of motor circuits may take significantly longer than the convergence of visual circuits, particularly if the figure is distant enough to be taken in with a single fixation. This should be the case because the visual mappings can be resolved with a partially solved kinematic template.

A number of psychophysical predictions also seem to be implied by this hypothesis. If the same kinematic circuit is used for planning and reconstruction, then it is likely that attempting to engage it concurrently for independent visual reconstruction and motor planning tasks should interfere with one or the other or both. Another psychophysical prediction is implied by the convergence behavior of the model. Due to the greater ambiguity in depth when stereo, perspectivity, shading and texture cues are weak or absent, the last surviving solutions tend to group tightly in the depth axis when the joint angle and other constraints aren't strong enough to separate them. An example is seen in Figure 17b, where a group of similar solutions is distributed perpendicular to the view plane. If this is also true in the brain, backward masking may leave convergence incomplete and with a multiplicity of tightly grouped solutions as described. In this case, the ability to precisely report or mimic a pose may be inhibited by backward masking at the right latency. (Alternatively, the biological system may pick a winner as soon as it is deprived of the relevant visual input, so this effect may be weak or absent.)

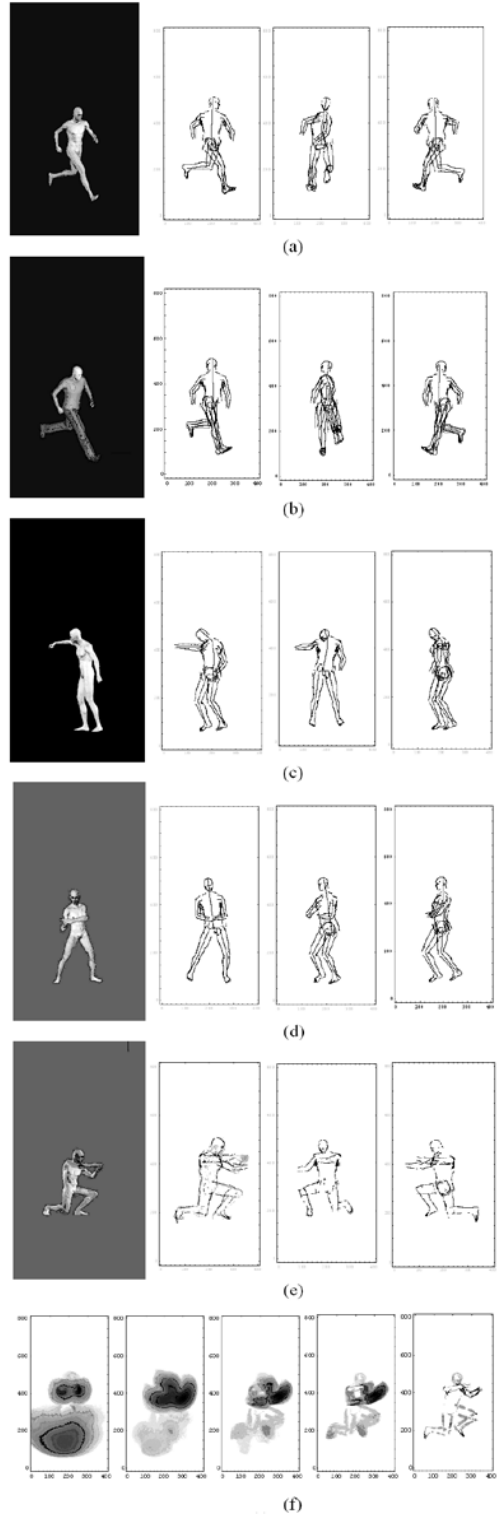

Figure 17. Example of pose reconstruction from computer graphics figures. (a,b) Two different morphs of the same pose. Note in (b) clustered solutions perpendicular to the image plane. (f) Convergence on the backward path (last frame is iteration 31). Note, (e) and (f) are from different morphing parameters using the same input image.

An hypothesis proposing function-specific computational circuits naturally raises the question of a more targeted assignment to the known anatomy. A number of imaging results attempt to situate our ability to recognize poses and action rendered in much reduced form, such as stick figures or joint markers, but leave the assignment to brain region ambiguous. For example, in [42] this capability is attributed to an area of LOC, while in [46] the premotor cortex is implicated. This uncertainty is not atypical, so anatomical assignment remains not much more than a guess. However, there are a few interesting area-specific experimental results which may be interpreted in light of the architecture presented here. For example, the reported presence of object category information of peripherally presented stimuli in the foveal area of V1 [41] might be interpreted as evidence that the peripheral presentation has caused a planned saccade to shift the target of interest to the central field. If the neuronal circuits involved have MSC-like behavior, then it is advantageous to force the $g$ coefficients of the translation stage to also shift the backward path projection of the just selected model and its various transformations to a small area of the fovea ahead of the the executed saccade. This means that there would be negligible convergence time for the centered image because all the $g$'s for all layers except the translation layer have already been resolved, and the translation layer needs only to select among a very few neighboring translations. This mode of operation of an MSC circuit for visual transformation is akin to the forward kinematic mode mentioned earlier. It is possible because MSC will function for both forward and inverse problems. In all cases it is advantageous to prime the backward path for the predicted target and, additionally, to restrict the transformations to just those which can be predicted to be necessary. This result is mentioned as an example of how the model proposed here may add to the interpretation of, or be amplified by, experiment in vivo.

## Acknowledgments

Simulations were supported by funding from Air Force Research Laboratory Sensors Directorate.